\documentclass[10pt,twocolumn,letterpaper]{article}

\usepackage{cvpr}
\usepackage{times}
\usepackage{epsfig}
\usepackage{graphicx}
\usepackage{amsmath}
\usepackage{amssymb}
\usepackage{mathrsfs}
\usepackage{slashbox}
\usepackage{dsfont}
\usepackage{booktabs}
\usepackage{subcaption}
\usepackage{caption}
\usepackage{nicefrac}
\usepackage{dblfloatfix}
\usepackage{mathrsfs}
\usepackage{xcolor}
\usepackage{enumitem}
\usepackage{colortbl}
\usepackage{multirow}

\usepackage{marvosym}
\usepackage[export]{adjustbox}
\usepackage{pifont}
\newcommand{\cmark}{\ding{51}}%
\newcommand{\xmark}{\ding{55}}%

\definecolor{sh_gray}{rgb}{0.84,0.84,0.84}
\definecolor{sh_gray2}{rgb}{1,0.89,0.75}
\definecolor{color3}{rgb}{0.95,0.95,0.95}
\definecolor{color4}{rgb}{0.96,0.96,0.86}
\definecolor{color5}{rgb}{0.90,0.90,0.90}

\newlength{\Oldarrayrulewidth}

\newcommand{\midsepremove}{\aboverulesep = 0mm \belowrulesep = 0mm}
\midsepremove
\newcommand{\midsepdefault}{\aboverulesep = 0.605mm \belowrulesep = 0.984mm}
\midsepdefault
  
\usepackage[pagebackref=true,breaklinks=true,letterpaper=true,colorlinks,linkcolor=blue,citecolor=blue,urlcolor=blue,bookmarks=false]{hyperref}

\cvprfinalcopy % *** Uncomment this line for the final submission

\usepackage{cuted}
\usepackage{capt-of}

\begin{document}
%%%%%%%%% TITLE
%%%%%%%%%%%%%%%%%%%%%%%%%%%%%%%%%%%%%%%%%%%%%%%%%%%%%%%%%%%%%%%%%%%%%%
\title{\vspace{-0.0em}Multi-Stage Progressive Image Restoration}

\author{
Syed Waqas Zamir\thanks{Equal contribution }~~$^{1}$ \quad Aditya Arora\footnotemark[1]~~$^{1}$  \quad Salman Khan$^{2}$ \quad Munawar Hayat$^{3}$ \\ 
Fahad Shahbaz Khan$^{2}$  \quad Ming-Hsuan Yang$^{4,5,6}$ \quad Ling Shao$^{1,2}$ \\
$^1$Inception Institute of AI \quad $^2$Mohamed bin Zayed University of AI \quad
$^3$Monash University\\
$^4$University of California, Merced \quad $^5$Yonsei University \quad $^6$Google Research 
\vspace{-0.5em}
}

\maketitle
 
\thispagestyle{empty}
% Pages are numbered in submission mode, and unnumbered in camera-ready
\ifcvprfinal\pagestyle{empty}\fi

% \vspace{-5em}
%%%%%%%%% ABSTRACT
\begin{abstract}\vspace{-0.85em}
Image restoration tasks demand a complex balance between spatial details and high-level contextualized information while recovering images. 
In this paper, we propose a novel synergistic design that can optimally balance these competing goals. Our main proposal is a multi-stage architecture, that progressively learns restoration functions for the degraded inputs, thereby breaking down the overall recovery process into more manageable steps. Specifically, our model first learns the contextualized features using encoder-decoder architectures and later combines them with a high-resolution branch that retains local information. At each stage, we introduce a novel per-pixel adaptive design that leverages in-situ supervised attention to reweight the local features. A key ingredient in such a multi-stage architecture is the information exchange between different stages. To this end, we propose a two-faceted approach where the information is not only exchanged sequentially from early to late stages, but lateral connections between feature processing blocks also exist to avoid any loss of information. The resulting tightly interlinked multi-stage architecture, named as MPRNet, delivers strong performance gains on ten datasets across a range of tasks including image deraining, deblurring, and denoising. %For example, on the Rain100L, GoPro and DND datasets, we obtain PSNR gains of $4$ dB, $0.81$ dB and $0.21$ dB, respectively, compared to the state-of-the-art. 
The source code and pre-trained models are available at \url{https://github.com/swz30/MPRNet}.  
\end{abstract}
\vspace{-0.3cm}
% %%------------------------------------

%%%%%%%%% BODY TEXT
\vspace{-0.5em}
\section{Introduction}
Image restoration is the task of recovering a clean image from its degraded version. Typical examples of degradation include noise, blur, rain, haze, etc.  
It is a highly ill-posed problem as there exist infinite feasible solutions.
In order to restrict the solution space to valid/natural images, existing restoration techniques~\cite{dong2011image,he2010single,kim2010single,perona1990scale,roth2005fields,rudin1992nonlinear,zhu1997prior} explicitly use image priors that are handcrafted with empirical observations.
However, designing such priors is a challenging task and often not generalizable.
To ameliorate this issue, recent state-of-the-art approaches~\cite{dai2019second,SRResNet,pan2020exploiting,zamir2020cycleisp,zamir2020mirnet,DnCNN,zhang2017learning,zhang2020rdn} employ convolutional neural networks (CNNs) that implicitly learn more general priors by capturing natural image statistics from large-scale data. 

The performance gain of CNN-based methods over the others is primarily attributed to its model design. 
Numerous network modules and functional units for image restoration have been developed including recursive residual learning~\cite{RIDNet,RCAN}, dilated convolutions~\cite{RIDNet,yang2017deep}, attention mechanisms~\cite{dai2019second,zamir2020cycleisp,zhang2019residual}, dense connections~\cite{tong2017image,wang2018esrgan,zhang2020rdn}, encoder-decoders~\cite{Brooks2019,Chen2018, deblurganv2,ronneberger2015unet}, and generative models~\cite{SRResNet,qian2018attentive,zhang2019image,zhang2020dbgan}.
Nevertheless, nearly all of these models for low-level vision problems are based on \emph{single-stage} design.  
In contrast, \emph{multi-stage} networks are shown to be more effective than their single-stage counterparts in high-level vision problems such as pose-estimation~\cite{chen2018cascaded,li2019rethinking,newell2016stacked}, scene parsing~\cite{cheng2019spgnet} and action segmentation~\cite{farha2019ms,ghosh2020stacked,li2020ms}.

\begin{figure}[!t]
\centering \hspace{-4mm}
    \begin{subfigure}[t]{0.9\columnwidth}
    \begin{picture}(100,150)
    \put(0,0){\includegraphics[width=\textwidth]{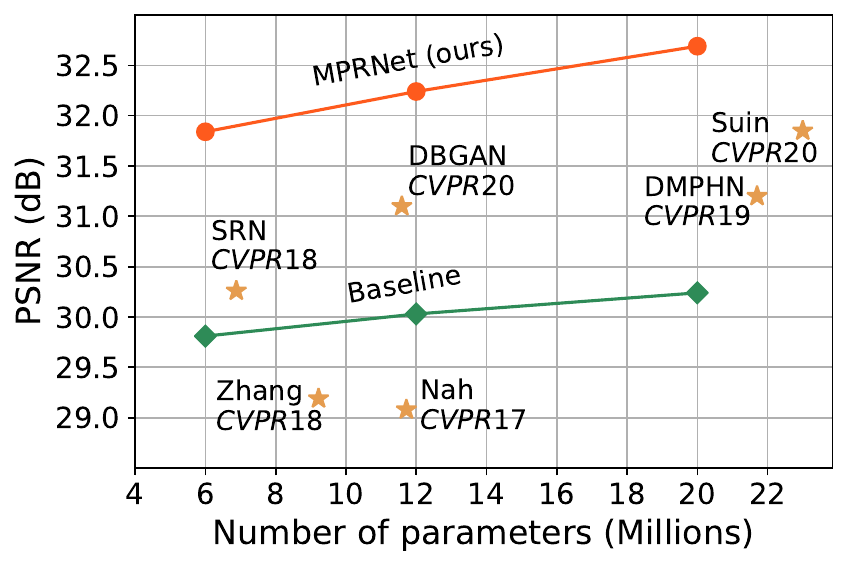}}
    \put(194,95){\scriptsize\cite{Maitreya2020}}
    \put(120,43){\scriptsize\cite{gopro2017}}
    \put(112,87){\scriptsize\cite{zhang2020dbgan}}
    \put(173,80){\scriptsize\cite{dmphn2019}}
    \put(60,28){\scriptsize\cite{zhang2018dynamic}}
    \put(70,83){\scriptsize\cite{tao2018scale}}
    \end{picture}
    \end{subfigure}
\vspace{-3mm}
\caption{\small Image deblurring on the GoPro dataset~\cite{gopro2017}. Under different parameter capacities (x-axis), our multi-stage approach performs better than the single-stage baseline~\cite{ronneberger2015unet} (with channel attention~\cite{RCAN}), as well as the state-of-the-art (PSNR  on y-axis). }\label{fig:intro}
\vspace{-1.5em}
\end{figure}
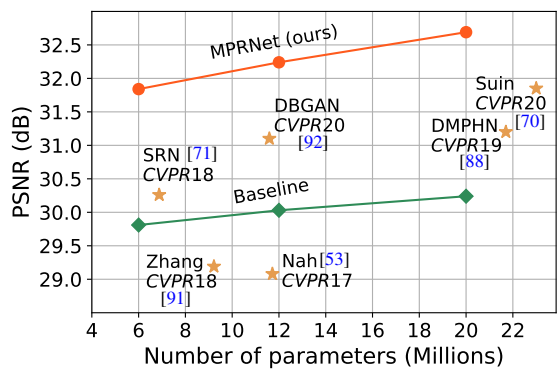

Recently, few efforts have been made to bring the multi-stage design to image deblurring~\cite{Maitreya2020,tao2018scale,dmphn2019}, and image deraining~\cite{li2018recurrent,ren2019progressive}.
We analyze these approaches to identify the architectural bottlenecks that hamper their performance.
First, existing multi-stage techniques either employ the \emph{encoder-decoder} architecture~\cite{tao2018scale,dmphn2019} which is effective in encoding broad contextual information but unreliable in preserving spatial image details, or use a \emph{single-scale pipeline}~\cite{ren2019progressive} that provides spatially accurate but semantically less reliable outputs.
However, we show that the combination of both design choices in a multi-stage architecture is needed for effective image restoration. 
Second, we show that naively passing the output of one stage to the next stage yields suboptimal results~\cite{gopro2017}. 
Third, unlike in~\cite{dmphn2019}, it is important to provide ground-truth supervision at each stage for progressive restoration.  
Finally, during multi-stage processing, a mechanism to propagate intermediate features from earlier to later stages is required to preserve contextualized features from the encoder-decoder branches.

We propose a multi-stage progressive image restoration architecture, called MPRNet, with several key components.
1). The earlier stages employ an encoder-decoder for learning multi-scale contextual information, while the last stage operates on the original image resolution to preserve fine spatial details. 
2). A supervised attention module (SAM) is plugged between every two stages to enable progressive learning. 
With the guidance of ground-truth image, this module exploits the previous stage prediction to compute attention maps that are in turn used to refine the previous stage features before being passed to the next stage.  
3). A mechanism of cross-stage feature fusion (CSFF) is added that helps propagating multi-scale contextualized features from the earlier to later stages. 
Furthermore, this method eases the information flow among stages, which is effective in stabilizing the multi-stage network optimization. 

\noindent The main contributions of this work are:\vspace{-0.5em}
\begin{itemize}[leftmargin=*]\setlength{\itemsep}{0em}
    \item  A novel multi-stage approach capable of generating contextually-enriched and spatially accurate outputs. Due to its multi-stage nature, our framework breaks down the challenging image restoration task into sub-tasks to progressively restore a degraded image. 
    \item An effective supervised attention module that takes full advantage of the restored image at every stage in refining incoming features before propagating them further. 
    \item A strategy to aggregate multi-scale features across stages. 
    \item We demonstrate the effectiveness of our MPRNet by setting new state-of-the-art on \textbf{ten} synthetic and real-world datasets for various restoration tasks including image deraining, deblurring, and denoising while maintaining a low complexity (see Fig.~\ref{fig:intro}). Further, we provide detailed ablations, qualitative results, and generalization tests. 
\end{itemize}

%%%%%%%%%%%%%%%%%%%%%%%%%%%%%%%%%%%%%%%%%%%%%%%%%%%%%%%%%%%%%%%%%%%%%%%%%%%%%%%
%%%%%%%%%%%%%%%%%%%%%%%%%%%%%%%%%%%%%%%%%%%%%%%%%%%%%%%%%%%%%%%%%%%%%%%%%%%%%%%
%%%%%%%%%%%%%%%%%%%%%%%%%%%%%%%%%%%%%%%%%%%%%%%%%%%%%%%%%%%%%%%%%%%%%%%%%%%%%%%
\section{Related Work}
Recent years have witnessed a paradigm shift from high-end DSLR cameras to smartphone cameras. 
However, capturing high-quality images with smartphone cameras is challenging. 
Image degradations are often present in images either due to the limitations of cameras and/or adverse ambient conditions. 
Early restoration approaches are based on total variation~\cite{chan1998total,rudin1992nonlinear}, sparse coding~\cite{KSVD,luo2015removing,mairal2007sparse}, self-similarity~\cite{NLM,BM3D}, gradient prior~\cite{shan2008high,xu2013unnatural}, \textit{etc}.
Recently, CNN-based restoration methods have achieved state-of-the-art results~\cite{pan2020exploiting,Maitreya2020,zamir2020cycleisp,DnCNN,zhang2020rdn}. 
In terms of architectural design, these methods can be broadly categorized as single-stage and multi-stage. 

%%%%%%%%%%%%%%%%%%%%%%%%%%%%%%%%%%%%%%%%%%%%%%%%%%%%%
\vspace{0.4em}\noindent\textbf{Single-Stage Approaches.} Currently, the majority of image restoration methods are based on a single-stage design, and the architectural components are usually based on those developed for high-level vision tasks. 
For example, residual learning~\cite{he2016deep} has been used to perform image denoising~\cite{ntire2019_denoising,tian2020deep,DnCNN}, image deblurring~\cite{deblurgan,deblurganv2} and image deraining~\cite{mspfn2020}. 
Similarly, to extract multi-scale information, the encoder-decoder~\cite{ronneberger2015unet} and dilated convolution~\cite{yu2015multi} models are often used~\cite{RIDNet,CBDNet,deblurganv2}.
Other single-stage approaches~\cite{anwar2020densely,zhang2018density,zhang2020rdn} incorporate dense connections~\cite{huang2017densely}. 
%

%%%%%%%%%%%%%%%%%%%%%%%%%%%%%%%%%%%%%%%%%%%%%%%%%%%%%
\vspace{0.4em}\noindent\textbf{Multi-Stage Approaches.} These methods~\cite{fu2019lightweight,li2018recurrent,gopro2017,ren2019progressive,Maitreya2020,tao2018scale,dmphn2019,zheng2019residual} aim to recover clean image in a progressive manner by employing a light-weight subnetwork at each stage.
Such a design is effective since it decomposes the challenging image restoration task into smaller easier sub-tasks.
However, a common practice is to use the identical subnetwork for each stage which may yield suboptimal results, as shown in our experiments (Section~\ref{sec: experiments}). 

\vspace{0.4em}\noindent\textbf{Attention.} Driven by its success in high-level tasks such as image classification~\cite{hu2018gather,hu2019squeeze,woo2018cbam}, segmentation~\cite{fu2019dual,huang2019ccnet} and detection~\cite{wang2018non,woo2018cbam}, attention modules have been used in low-level vision tasks~\cite{khan2021transformers}. 
Examples abound, including methods for image deraining~\cite{mspfn2020,li2018recurrent}, deblurring~\cite{purohit2020region,Maitreya2020}, super-resolution~\cite{dai2019second,RCAN}, and denoising~\cite{RIDNet,zamir2020cycleisp}. 
The main idea is to capture long-range inter-dependencies along spatial dimensions~\cite{zhao2018psanet}, channel dimensions~\cite{hu2019squeeze}, or both~\cite{woo2018cbam}.

%%%%%%%%%%%%%%%%%%%%%%%%%%%%%%%%%%%%%%%%%%%%%%%%%%%%%%%%%%%%%%%%%%%%%%%%%%%%%%%
%%%%%%%%%%%%%%%%%%%%%%%%%%%%%%%%%%%%%%%%%%%%%%%%%%%%%%%%%%%%%%%%%%%%%%%%%%%%%%%%%%%%%%%%%%%%%%%%%%%%%%%%%%%%%%%%%%%%%%%%%%%%%%%%%%%%%%%%%%%%%%%%%%%%%%%%%%%%%%
\section{Multi-Stage Progressive Restoration}
\label{sec:method}
\begin{figure}[t!]
\begin{center}
 \includegraphics[width=\linewidth]{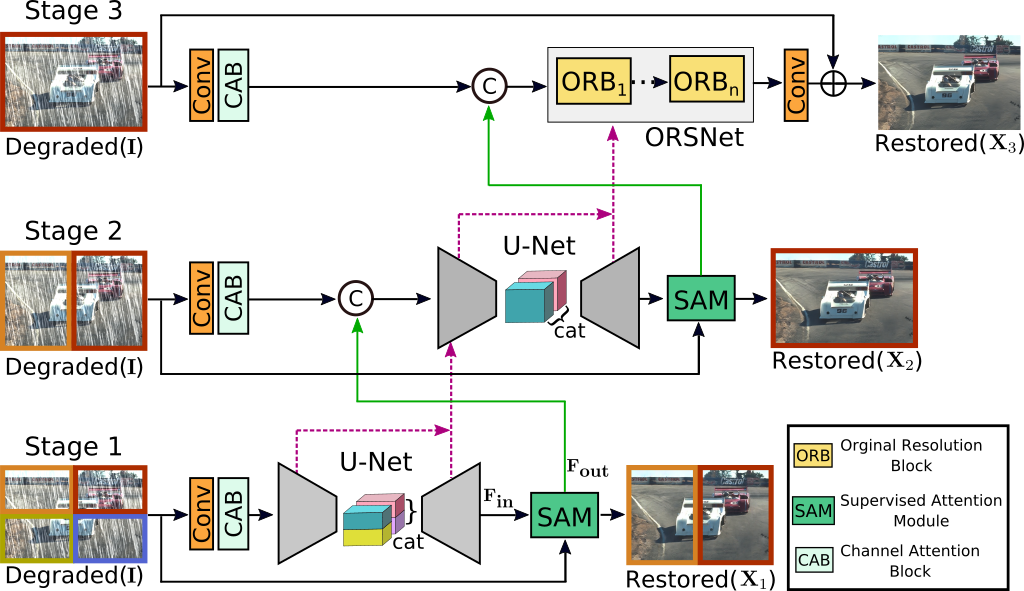}  
\end{center}\vspace{-1.4em}
    \caption{ \small 
    Proposed multi-stage architecture for progressive image restoration. Earlier stages employ encoder-decoders to extract multi-scale contextualized features, while the last stage operates at the original image resolution to generate spatially accurate outputs.    
    A supervised attention module is added between every two stages that learns to refine features of one stage before passing them to the next stage. Dotted pink arrows represent the cross-stage feature fusion mechanism. }
    \label{fig:framework}
\vspace{-1em}
\end{figure}

The proposed framework for image restoration, shown in Fig.~\ref{fig:framework}, consists of three stages to progressively restore images.
The first two stages are based on encoder-decoder subnetworks that learn the broad contextual information due to large receptive fields.
Since image restoration is a position-sensitive task (which requires pixel-to-pixel correspondence from the input to output), the last stage employs a subnetwork that operates on the original input image resolution (without any downsampling operation), thereby preserving the desired fine texture in the final output image.  

Instead of simply cascading multiple stages, we incorporate a supervised attention module between every two stages. 
With the supervision of ground-truth images, our module rescales the feature maps of the previous stage before passing them to the next stage. 
Furthermore, we introduce a cross-stage feature fusion mechanism where the intermediate multi-scale contextualized features of the earlier subnetwork help consolidating the intermediate features of the latter subnetwork. 

Although MPRNet stacks multiple stages, each stage has an access to the input image. 
Similar to the recent restoration methods~\cite{Maitreya2020,dmphn2019}, we adapt the multi-patch hierarchy on the input image and split the image into non-overlapping patches: four for stage-$1$, two for stage-$2$, and the original image for the last stage, as shown in Fig.~\ref{fig:framework}.    

\begin{figure*}[!t]
\centering
    \begin{subfigure}[t]{0.3\textwidth}
      \includegraphics[width=\textwidth]{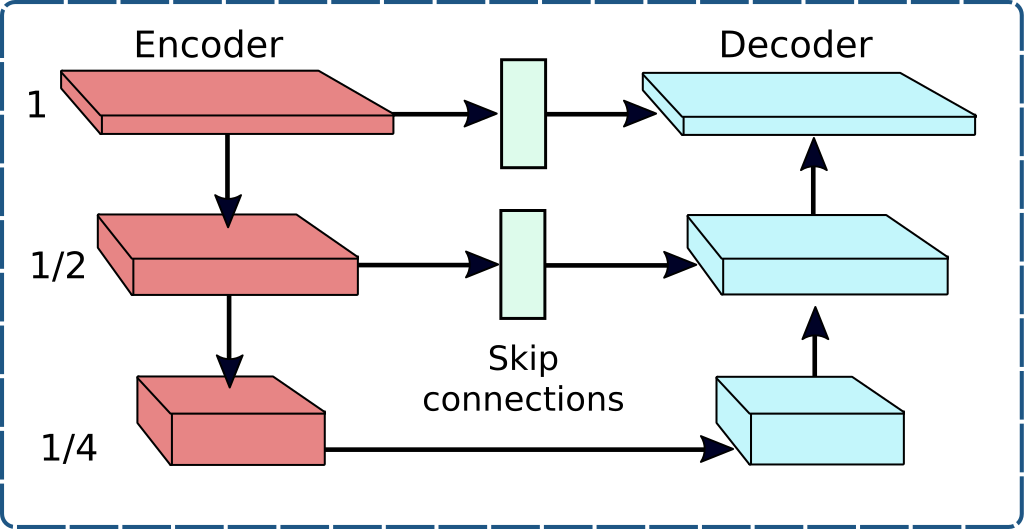}
      \caption{\small}
      \label{fig:unet}
    \end{subfigure}
    \begin{subfigure}[t]{0.35\textwidth}
      \includegraphics[width=\textwidth]{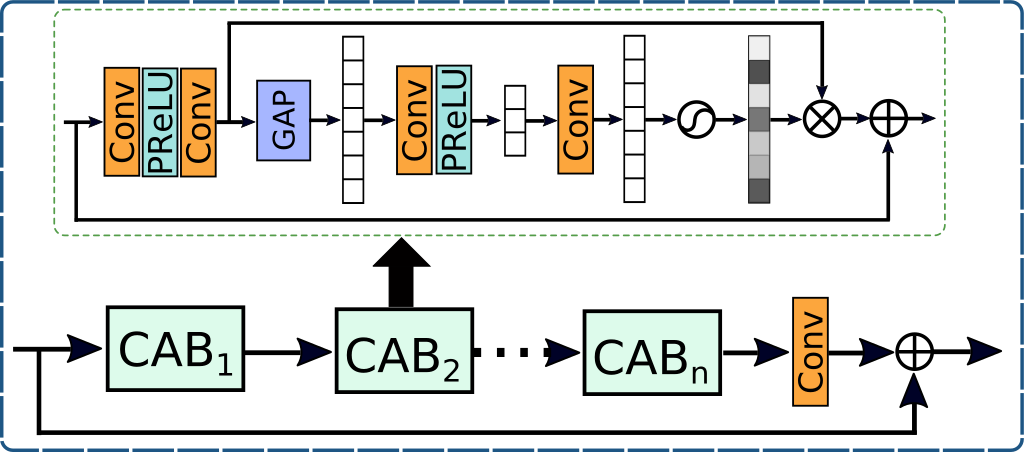}
      \caption{\small }
      \label{fig:orb}
    \end{subfigure}
    \begin{subfigure}[t]{0.155\textwidth}
      \includegraphics[width=\textwidth]{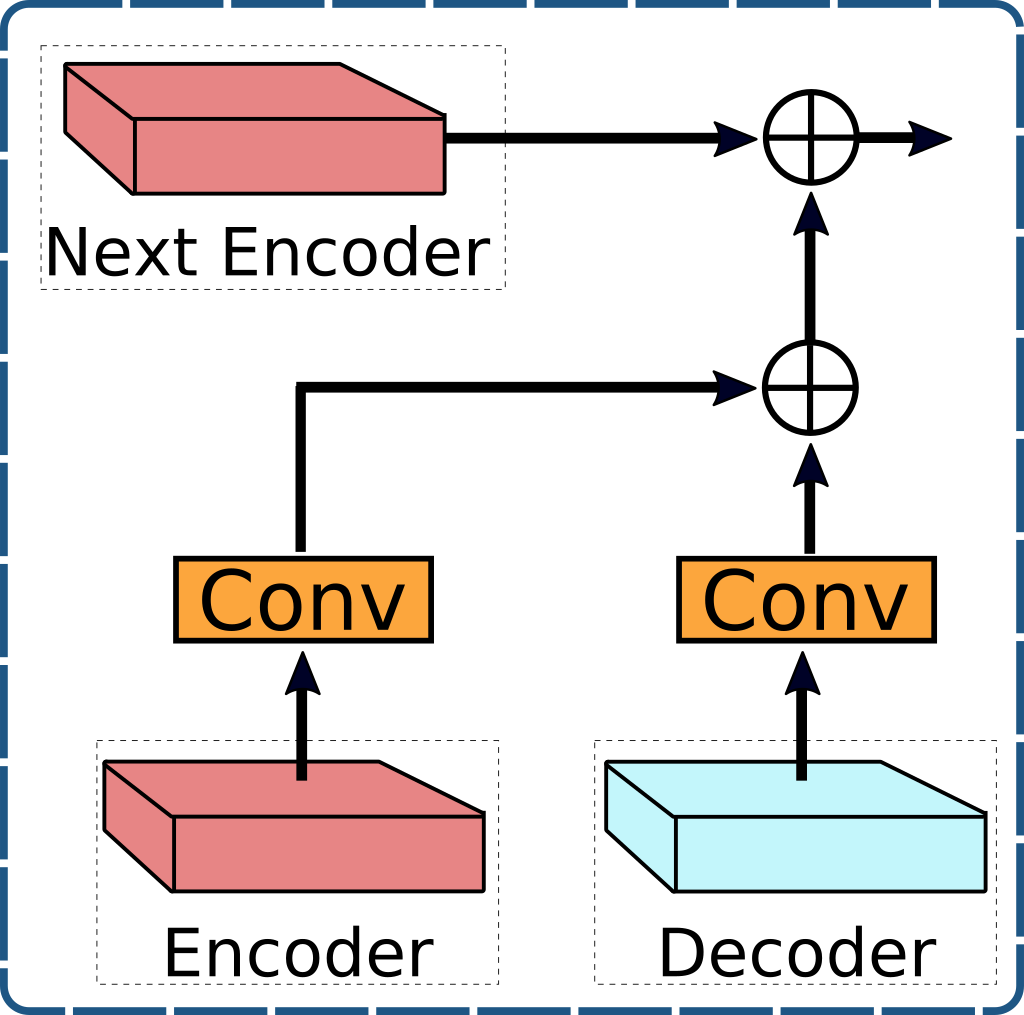}
      \caption{\small }
      \label{fig:csff unets}
    \end{subfigure}
    \begin{subfigure}[t]{0.155\textwidth}
      \includegraphics[width=\textwidth]{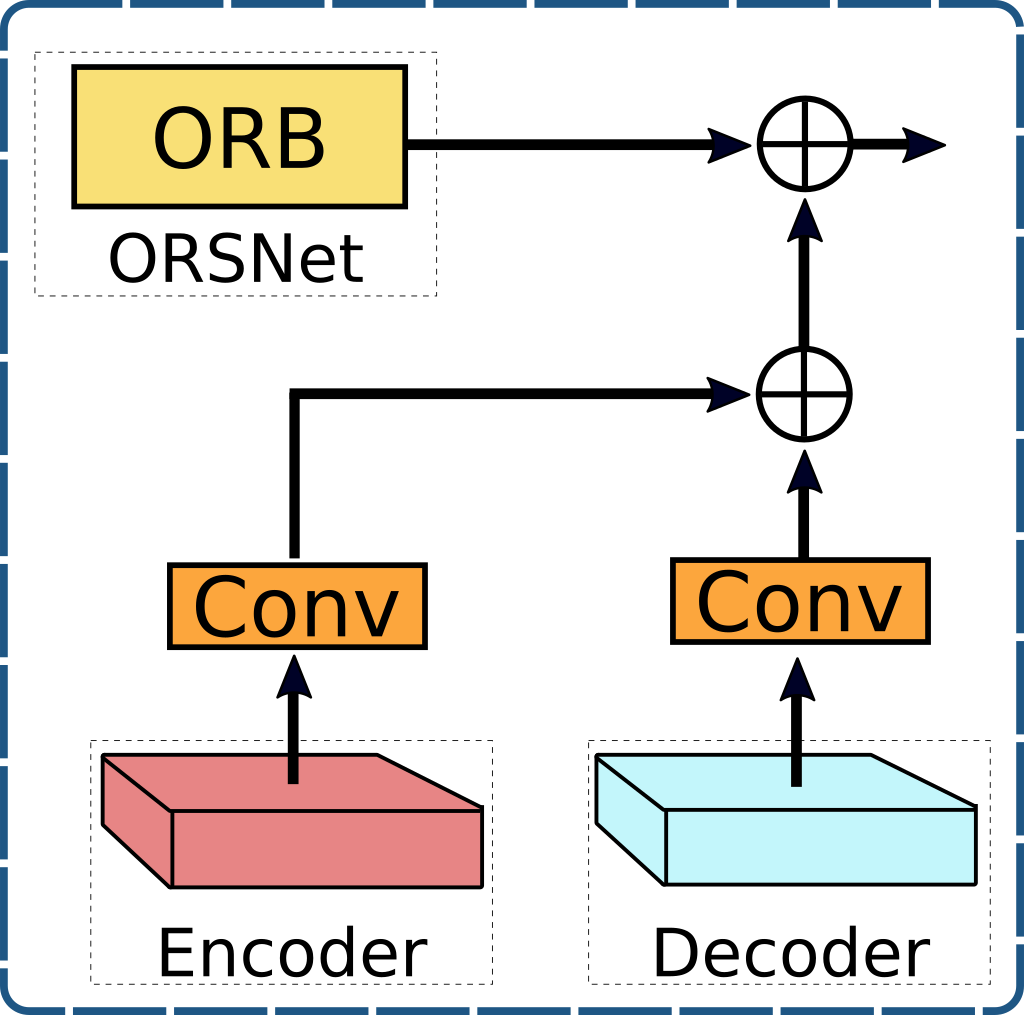}
      \caption{\small}
      \label{fig:csff orsnet}
    \end{subfigure}
\vspace{-3mm}
\caption{\small (a) Encoder-decoder subnetwork. (b) Illustration of the original resolution block (ORB) in our ORSNet subnetwork. Each ORB contains multiple channel attention blocks. GAP represents global average pooling~\cite{liu2015gap}. (c) Cross-stage feature fusion between stage 1 and stage 2. (d) CSFF between stage 2 and the last stage. }
\vspace{-1.5em}
\end{figure*}

At any given stage ${S}$, instead of directly predicting a restored image $\mathbf{X}_S$, the proposed model predicts a residual image $\mathbf{R}_S$ to which the degraded input image $\mathbf{I}$ is added to obtain: $\mathbf{X}_S = \mathbf{I} + \mathbf{R}_S$.
We optimize our MPRNet end-to-end with the following loss function:
\begin{equation}
\label{Eq:loss}
\mathcal{L} = \sum_{S=1}^{3}  \left[ \mathcal{L}_{char}(\mathbf{X}_S,\mathbf{Y}) + \lambda \mathcal{L}_{edge}(\mathbf{X}_S,\mathbf{Y}) \right],
\end{equation}
where $\mathbf{{Y}}$ represents the ground-truth image, and  $\mathcal{L}_{char}$ is the Charbonnier loss \cite{charbonnier1994}: 
\begin{equation}
\label{Eq:charbonnier}
\mathcal{L}_{char} = \sqrt{ {\|\mathbf{X}_S-\mathbf{Y}\|}^2 + {\varepsilon}^2 },
\end{equation}
with constant $\varepsilon$ empirically set to $10^{-3}$ for all the experiments. 
In addition, $\mathcal{L}_{edge}$ is the edge loss, defined as: 
\begin{equation}
\label{Eq:edge}
\mathcal{L}_{edge} = \sqrt{ {\|\Delta (\mathbf{X}_S) - \Delta (\mathbf{Y})\|}^2 + {\varepsilon}^2 },
\end{equation}
where $\Delta$ denotes the Laplacian operator. The parameter $\lambda$ in Eq.~(\ref{Eq:loss}) controls the relative importance of the two loss terms, which is set to $0.05$ as in~\cite{mspfn2020}. 
Next, we describe each key element of our method. 

%%%%%%%%%%%%%%%%%%%%%%%%%%%%%%%%%%%%%%%%%%%%%%%%%%%%%%%%%%%%%%%%%%%%%%%%%%%%%%%
\subsection{Complementary Feature Processing}

Existing single-stage CNNs for image restoration typically use one of the following architecture designs: 1). An encoder-decoder, or 2). A single-scale feature pipeline. 
The encoder-decoder networks \cite{Brooks2019,Chen2018, deblurganv2,ronneberger2015unet} first gradually map the input to low-resolution representations, and then progressively apply reverse mapping to recover the original resolution. 
While these models effectively encode multi-scale information, they are prone to sacrificing spatial details due to the repeated use of downsampling operation. 
In contrast, the approaches that operate on single-scale feature pipeline are reliable in generating images with fine spatial details~\cite{anwar2019deep,dong2015image,DnCNN,zhang2020rdn}.
However, their outputs are semantically less robust due to the limited receptive field.  
This indicates the inherent limitations of the aforementioned architecture design choices that are capable of generating either spatially accurate or contextually reliable outputs, but not both. 
To exploit the merits of both designs, we propose a multi-stage framework where earlier stages incorporate the encoder-decoder networks, and the final stage employs a network that operates on the original input resolution. 

%%%%%%%%%%%%%%%%%%%%%%%%%%%%%%%%%%%%%%%%%%%%%%%%%%%%%%%%%%%%%%%%%%%%%%%%%%%%%%%
\vspace{0.4em}\noindent\textbf{Encoder-Decoder Subnetwork.}
Figure~\ref{fig:unet} shows our encoder-decoder subnetwork, which is based on the standard U-Net~\cite{ronneberger2015unet}, with the following components. 
First, we add channel attention blocks (CABs)~\cite{RCAN} to extract features at each scale (See Fig.~\ref{fig:orb} for CABs).
Second, the feature maps at U-Net skip connections are also processed with the CAB.
Finally, instead of using Transposed convolution for increasing spatial resolution of features in the decoder, we use bilinear upsampling followed by a convolution layer. 
This helps reduce checkerboard artifacts in the output image that often arise due to the Transposed convolution~\cite{Odena2016}.  
%

%%%%%%%%%%%%%%%%%%%%%%%%%%%%%%%%%%%%%%%%%%%%%%%%%%%%%%%%%%%%%%%%%%%%%%%%%%%%%%%
\vspace{0.4em}\noindent\textbf{Original Resolution Subnetwork.}
In order to preserve fine details from the input image to the output image, we introduce the original-resolution subnetwork (ORSNet) in the last stage (see Fig.~\ref{fig:framework}). ORSNet does not employ any downsampling operation and generates spatially-enriched high-resolution features. It consists of multiple original-resolution blocks (ORBs), each of which further contains CABs.  The schematic of ORB is illustrated in Fig.~\ref{fig:orb}.

\begin{figure}[t]
\begin{center}
 \includegraphics[width=\linewidth]{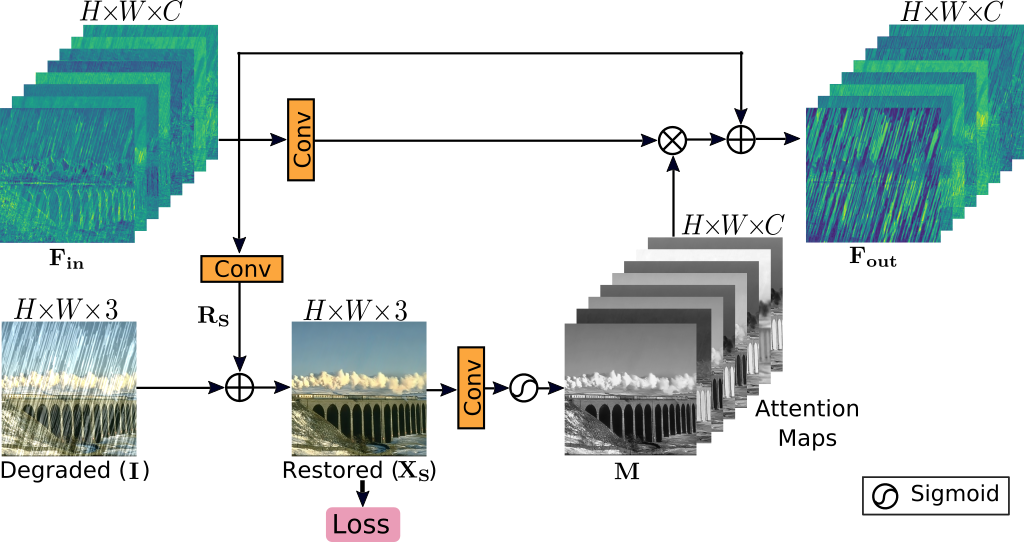}  
\end{center}\vspace{-1.4em}
    \caption{ \small Supervised attention module.}
    \vspace{-0.5em}
    \label{fig:sam}
\vspace{-1.4em}
\end{figure}

%%%%%%%%%%%%%%%%%%%%%%%%%%%%%%%%%%%%%%%%%%%%%%%%%%%%%%%%%%%%%%%%%%%%%%%%%%%%%%%
\subsection{Cross-stage Feature Fusion}

In our framework, we introduce the CSFF module between two encoder-decoders (see Fig.~\ref{fig:csff unets}), and between encoder-decoder and ORSNet (see Fig.~\ref{fig:csff orsnet}).
Note that the features from one stage are first refined with $1\times1$ convolutions before propagating them to the next stage for aggregation. 
The proposed CSFF has several merits. 
First, it makes the network less vulnerable by the information loss due to repeated use of up- and down-sampling operations in the encoder-decoder. 
Second, the multi-scale features of one stage help enriching the features of the next stage. 
Third, the network optimization procedure becomes more stable as it eases the flow of information, thereby allowing us to add several stages in the overall architecture. 
%

%%%%%%%%%%%%%%%%%%%%%%%%%%%%%%%%%%%%%%%%%%%%%%%%%%%%%%%%%%%%%%%%%%%%%%%%%%%%%%%

\subsection{Supervised Attention Module}
Recent multi-stage networks for image restoration~\cite{Maitreya2020,dmphn2019} directly predict an image at each stage, which is then passed to the next consecutive stage.
Instead, we introduce a supervised attention module between every two stages, which facilitates achieving significant performance gain.
The schematic diagram of SAM is shown in Fig.~\ref{fig:sam}, and its contributions are two-fold. 
First, it provides ground-truth supervisory signals useful for the progressive image restoration at each stage. 
Second, with the help of locally supervised predictions, we generate attention maps to suppress the less informative features at the current stage and only allow the useful ones to propagate to the next stage.    

As illustrated in Fig.~\ref{fig:sam}, SAM takes the incoming features $\mathbf{F_{in}} \in \mathbb{R}^{H\times W \times C}$ of the earlier stage and first generates a residual image $\mathbf{R}_S \in \mathbb{R}^{H\times W \times 3}$ with a simple $1\times1$ convolution, where $H\times W$ denotes the spatial dimension and $C$ is the number of channels. 
The residual image is added to the degraded input image $\mathbf{I}$ to obtain the restored image $\mathbf{X}_S \in \mathbb{R}^{H\times W \times 3}$.
To this predicted image $\mathbf{X}_S$, we provide explicit supervision with the ground-truth image. 
Next, per-pixel attention masks $\mathbf{M} \in \mathbb{R}^{H\times W \times C}$ are generated from the image $\mathbf{X}_S$ using a $1\times1$ convolution followed by the sigmoid activation. 
These masks are then used to re-calibrate the transformed local features $\mathbf{F_{in}}$ (obtained after $1\times1$ convolution), resulting in attention-guided features which are added to the identity mapping path. 
Finally, the attention-augmented feature representation $\mathbf{F_{out}}$, produced by SAM, is passed to the next stage for further processing. 

%%%%%%%%%%%%%%%%%%%%%%%%%%%%%%%%%%%%%%%%%%%%%%%%%%%%%%%%%%%%%%%%%%%%

\begin{table*}[!t]
\begin{center}
\caption{\small Dataset description for various image restoration tasks.}
\label{table:datasets}
\vspace{-2mm}
\setlength{\tabcolsep}{0.8pt}
\scalebox{0.70}{
\begin{tabular}{l c c c c c c c | c c c | c c }
\toprule[0.15em]
 Tasks & \multicolumn{7}{c|}{Deraining} & \multicolumn{3}{c|}{Deblurring} & \multicolumn{2}{c}{Denoising}\\
\toprule[0.15em]
Datasets  &   Rain14000~\cite{fu2017removing}  &  Rain1800~\cite{yang2017deep}  & Rain800~\cite{zhang2019image} & Rain100H~\cite{yang2017deep} & Rain100L~\cite{yang2017deep} & Rain1200~\cite{zhang2018density} & Rain12~\cite{li2016rain} & GoPro~\cite{gopro2017}  & HIDE~\cite{shen2019human} & RealBlur~\cite{rim_2020_realblur} & SIDD~\cite{sidd}  & DND~\cite{dnd} \\
Train Samples     &   11200      &  1800      & 700     & 0        & 0        & 0        & 12     & 2103   & 0  & 0  & 320   & 0   \\
Test Samples      &   2800       &  0         & 100     & 100      & 100      & 1200     & 0      & 1111   & 2025 & 1960 & 40  & 50 \\
\midrule
Testset Rename & Test2800 & - & Test100 &	Rain100H	& Rain100L & Test1200  & - & - & -& - & - & - \\
\bottomrule[0.1em]
\end{tabular}}
\end{center}\vspace{-1.5em}
\end{table*}

%%%%%%%%%%%%%%%%%%%%%%%%%%%%%%%%%%%%%%%%%%%%%%%%%%%%%%%%%%%%%%%%%%%%
%%%%%%%%%%%%%%%%%%%%%%%%%%%%%%%%%%%%%%%%%%%%%%%%%%%%%%%%%%%%%%%%%%%%
%%%%%%%%%%%%%%%%%%%%%%%%%%%%%%%%%%%%%%%%%%%%%%%%%%%%%%%%%%%%%%%%%%%%

\section{Experiments and Analysis} 
\label{sec: experiments}
We evaluate our method for several image restoration tasks, including \textbf{(a)} image deraining, \textbf{(b)} image deblurring, and \textbf{(c)} image denoising on $10$ different datasets.
%

%%%%%%%%%%%%%%%%%%%%%%%%%%%%%%%%%%%%%%%%%%%%%%%%%%%%%%%%%%%%%%%%%%%%

\subsection{Datasets and Evaluation Protocol}
\label{sec:datasets}
 Quantitative comparisons are performed using the PSNR and SSIM~\cite{Wang2004ssim} metrics.
 As in~\cite{Brooks2019}, we  report (in parenthesis) the reduction in error for each method relative to the best performing method by translating PSNR to RMSE ($\textrm{RMSE} \propto \sqrt{10^{-\textrm{PSNR}/10}}$) and SSIM to DSSIM ($\textrm{DSSIM} = (1 - \textrm{SSIM})/2$). 
The datasets used for training and testing are summarized in Table~\ref{table:datasets} and described next.

%%%%%%%%%%%%%%%%%%%%%%%%%%%%%%%%%%%%%%%%%%%%%%%%%%%%%%%%%%%%%%%%%%%%%%%%%%%%

\begin{table*}
\begin{center}
\caption{\small Image deraining results. Best and second best scores are \textbf{highlighted} and \underline{underlined}.  
For each method, reduction in error relative to the best-performing algorithm is reported in parenthesis (see Section~\ref{sec:datasets} for error calculation technique). 
Our MPRNet achieves $\sim$$20\%$ relative improvement in PSNR over the previous best method MSPFN~\cite{mspfn2020}. }
\label{table:deraining}
\vspace{-2mm}
\setlength{\tabcolsep}{5.9pt}
\scalebox{0.74}{
\begin{tabular}{l c c c c c c c c c c || c c}
\toprule[0.15em]
  & \multicolumn{2}{c}{Test100~\cite{zhang2019image}}&\multicolumn{2}{c}{Rain100H~\cite{yang2017deep}}&\multicolumn{2}{c}{Rain100L~\cite{yang2017deep}}&\multicolumn{2}{c}{Test2800~\cite{fu2017removing}}&\multicolumn{2}{c||}{Test1200~\cite{zhang2018density}}&\multicolumn{2}{c}{Average}\\
 Methods & PSNR~$\textcolor{black}{\uparrow}$ & SSIM~$\textcolor{black}{\uparrow}$ & PSNR~$\textcolor{black}{\uparrow}$ & SSIM~$\textcolor{black}{\uparrow}$ & PSNR~$\textcolor{black}{\uparrow}$ & SSIM~$\textcolor{black}{\uparrow}$ & PSNR~$\textcolor{black}{\uparrow}$ & SSIM~$\textcolor{black}{\uparrow}$ & PSNR~$\textcolor{black}{\uparrow}$ & SSIM~$\textcolor{black}{\uparrow}$ & PSNR~$\textcolor{black}{\uparrow}$ & SSIM~$\textcolor{black}{\uparrow}$\\
\midrule[0.15em]
DerainNet~\cite{fu2017clearing} & 22.77  & 0.810  & 14.92  & 0.592  & 27.03  & 0.884  & 24.31  & 0.861  & 23.38  & 0.835  & 22.48 \colorbox{gray!20}{(69.3\%)}  & 0.796  \colorbox{gray!20}{(61.3\%)}  \\
SEMI~\cite{wei2019semi} & 22.35&0.788& 16.56&0.486& 25.03&0.842& 24.43&0.782& 26.05&0.822& 22.88 \colorbox{gray!20}{(67.8\%)}&0.744 \colorbox{gray!20}{(69.1\%)}\\
DIDMDN~\cite{zhang2018density} & 22.56&0.818& 17.35&0.524& 25.23&0.741& 28.13&0.867& 29.65&0.901& 24.58 \colorbox{gray!20}{(60.9\%)}&0.770 \colorbox{gray!20}{(65.7\%)} \\
UMRL~\cite{yasarla2019uncertainty} & 24.41&0.829& 26.01&0.832& 29.18&0.923& 29.97&0.905& 30.55&0.910& 28.02 \colorbox{gray!20}{(41.9\%)}&0.880 \colorbox{gray!20}{(34.2\%)} \\
RESCAN~\cite{li2018recurrent} & 25.00&0.835& 26.36&0.786& 29.80&0.881& 31.29&0.904& 30.51&0.882& 28.59 \colorbox{gray!20}{(37.9\%)} & 0.857 \colorbox{gray!20}{(44.8\%)} \\
PreNet~\cite{ren2019progressive} & 24.81&0.851& 26.77&0.858& \underline{32.44} & \underline{0.950} & 31.75&0.916& 31.36&0.911& 29.42 \colorbox{gray!20}{(31.7\%)} &0.897 \colorbox{gray!20}{(23.3\%)}\\
MSPFN~\cite{mspfn2020}  & \underline{27.50} & \underline{0.876} & \underline{28.66} & \underline{0.860} & 32.40&0.933& \underline{32.82} & \underline{0.930} & 32.39 & \underline{0.916} & \underline{30.75} \colorbox{gray!20}{(20.4\%)} & \underline{0.903} \colorbox{gray!20}{(18.6\%)} \\
\midrule
\textbf{MPRNet (Ours)}  & \textbf{30.27} & \textbf{0.897} & \textbf{30.41} & \textbf{0.890} & \textbf{36.40} & \textbf{0.965} & \textbf{33.64} & \textbf{0.938} & \textbf{32.91} & \textbf{0.916} & \textbf{32.73} \colorbox{gray!20}{(0.0\%)}& \textbf{0.921} \colorbox{gray!20}{(0.0\%)}\\

\bottomrule[0.1em]
\end{tabular}}
\end{center}\vspace{-1.3em}
\end{table*}

%%%%%%%%%%%%%%%%%%%%%%%%%%%%%%%%%%%%%%%%%%%%%%%%%%%%%%%%%%%%%%%%%%%%%%%%%%%%%%%%%%%%%%%%%%%%%

\vspace{0.4em}\noindent\textbf{Image Deraining.}
Using the same experimental setups of the recent best method on image deraining~\cite{mspfn2020}, we train our model on $13$,$712$ clean-rain image pairs gathered from multiple datasets~\cite{fu2017removing,li2016rain,yang2017deep,zhang2018density,zhang2019image}, as shown in Table~\ref{table:datasets}. 
With this single trained model, we perform evaluation on various test sets, including Rain100H~\cite{yang2017deep}, Rain100L~\cite{yang2017deep}, Test100~\cite{zhang2019image}, Test2800~\cite{fu2017removing}, and Test1200~\cite{zhang2018density}.

%%%%%%%%%%%%%%%%%%%%%%%%%%%%%%%%%%%%%%%%%%%%%%%%%%%%%%%%%%%%%%%%%%%%%%%%%%%%%%%%%%%%%%%%%%%%%
\vspace{0.4em}\noindent\textbf{Image Deblurring.}
As in \cite{Maitreya2020, dmphn2019, deblurganv2, tao2018scale}, we use the GoPro \cite{gopro2017} dataset that contains $2$,$103$ image pairs for training and $1$,$111$ pairs for evaluation.
Furthermore, to demonstrate generalizability, we take our GoPro trained model and \emph{directly apply} it on the test images of the  HIDE~\cite{shen2019human} and RealBlur~\cite{rim_2020_realblur} datasets. 
The HIDE dataset is specifically collected for human-aware motion deblurring and its test set contains $2$,$025$ images. 
While the GoPro and HIDE datasets are synthetically generated, the image pairs of RealBlur dataset are captured in real-world conditions. 
The RealBlur dataset has two subsets: (1) RealBlur-J is formed with the camera JPEG outputs, and (2) RealBlur-R is generated offline by applying white balance, demosaicking, and denoising operations to the RAW images.

%%%%%%%%%%%%%%%%%%%%%%%%%%%%%%%%%%%%%%%%%%%%%%%%%%%%%%%%%%%%%%%%%%%%%%%%%%%%%%%%%%%%%%%%%%%%%

\vspace{0.4em}\noindent\textbf{Image Denoising.}
To train our model for image denoising task, we use $320$ high-resolution images of the SIDD dataset~\cite{sidd}. 
Evaluation is conducted on $1$,$280$ validation patches from the SIDD dataset~\cite{sidd} and $1$,$000$ patches from the DND benchmark dataset~\cite{dnd}. 
These test patches are extracted from the full resolution images by the original authors. 
Both SIDD and DND datasets consist of real images. 

\begin{figure*}[!t]
\begin{center}
\scalebox{0.97}{
\begin{tabular}[b]{c@{ } c@{ }  c@{ } c@{ } c@{ } c@{ }	}\hspace{-4mm}
    \multirow{4}{*}{\includegraphics[width=.326\textwidth,valign=t]{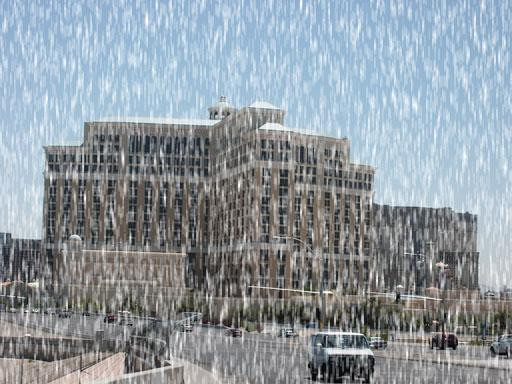}} &   
    \includegraphics[trim={ 192 87 250 244
 },clip,width=.13\textwidth,valign=t]{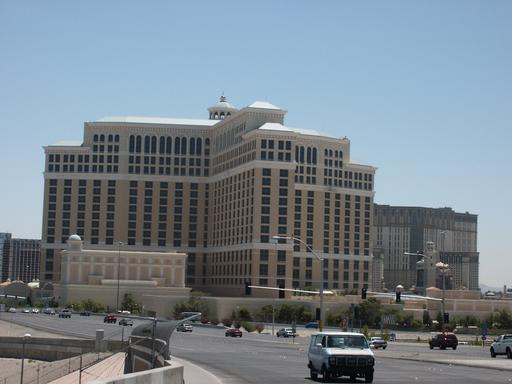}&
  	\includegraphics[trim={192 87 250 244 },clip,width=.13\textwidth,valign=t]{Images/Deraining/img1/input_18_76.jpg}&   
    \includegraphics[trim={192 87 250 244 },clip,width=.13\textwidth,valign=t]{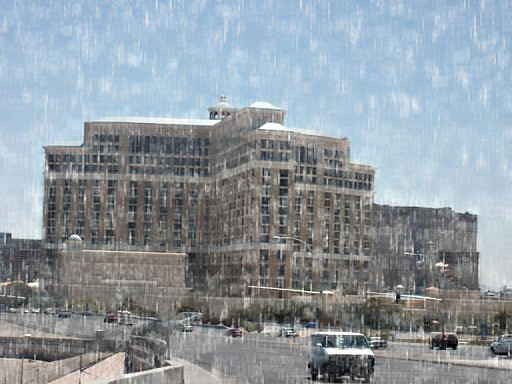}&
      	\includegraphics[trim={192 87 250 244 },clip,width=.13\textwidth,valign=t]{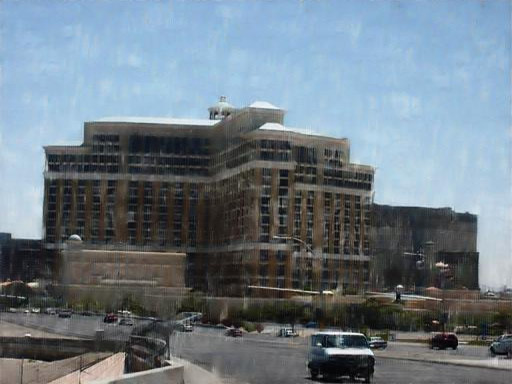}&
      \includegraphics[trim={192 87 250 244 },clip,width=.13\textwidth,valign=t]{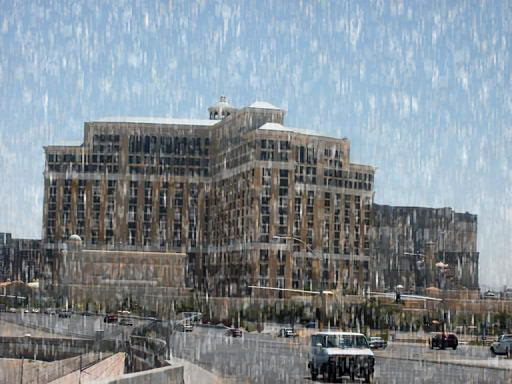}
  
\\
    &  \small~PSNR &\small~18.76 dB  & \small~20.23 dB & \small~23.36 dB & \small~23.66 dB   \\
    & \small~Reference & \small~Rainy  & \small~DerainNet~\cite{fu2017clearing}  & \small~DIDMDN~\cite{zhang2018density} & \small~SEMI~\cite{wei2019semi} \\

    &
    \includegraphics[trim={192 87 250 244 },clip,width=.13\textwidth,valign=t]{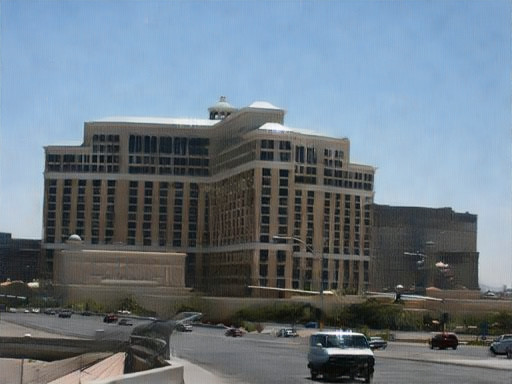}&
    \includegraphics[trim={192 87 250 244 },clip,width=.13\textwidth,valign=t]{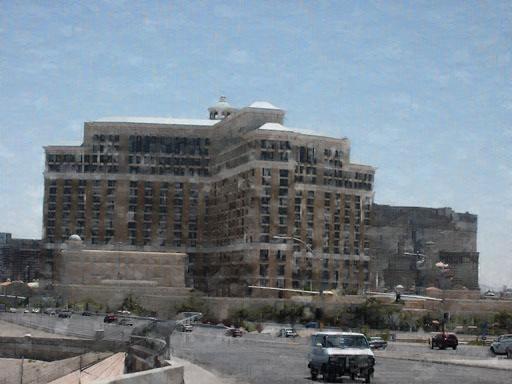}&
    \includegraphics[trim={192 87 250 244 },clip,width=.13\textwidth,valign=t]{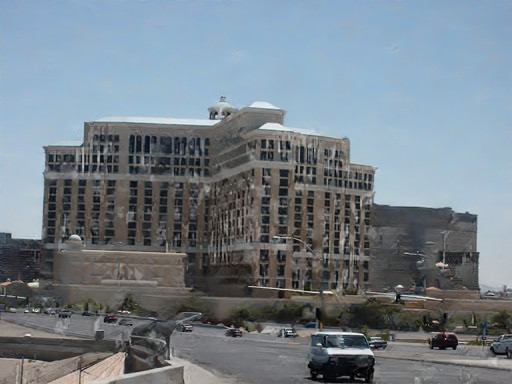}&  
     \includegraphics[trim={192 87 250 244 },clip,width=.13\textwidth,valign=t]{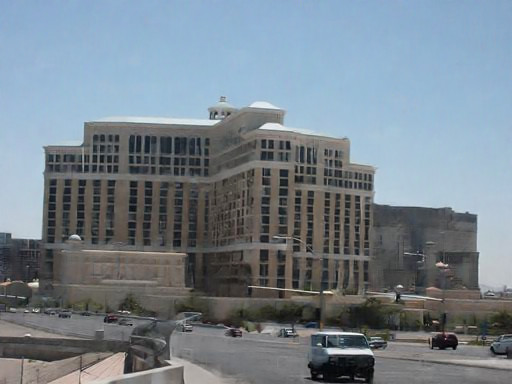}&
     \includegraphics[trim={192 87 250 244 },clip,width=.13\textwidth,valign=t]{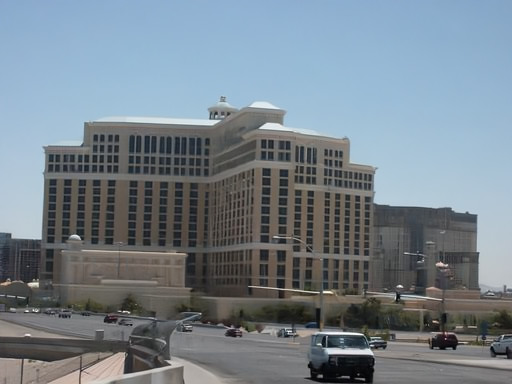}\\

     \small~18.76 dB& \small~25.52 dB& \small~26.88 dB & \small~27.16 dB 
     & \small~29.86 dB & \small~\textbf{32.15 dB}\\
           \small~Rainy Image  & \small~UMRL~\cite{yasarla2019uncertainty}& \small~RESCAN~\cite{li2018recurrent} & \small~PreNet~\cite{ren2019progressive} & \small~MSPFN~\cite{mspfn2020}   & \small~\textbf{MPRNet (Ours)}
\\
\end{tabular}}
\end{center}
\vspace{-6mm}
\end{figure*}

\begin{figure*}[!t]
\begin{center}
\scalebox{1}{
\begin{tabular}[t]{c@{ }c@{ }c@{ }c@{ }c@{ }c@{ }c@{ }c}\hspace{-4mm}
\includegraphics[width=.12\textwidth]{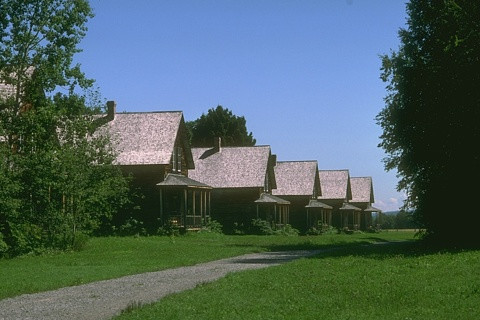}&   \hspace{-1.4mm}
\includegraphics[width=.12\textwidth]{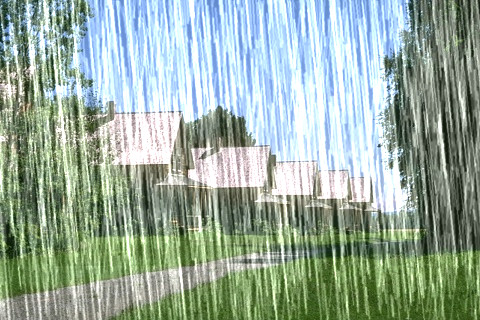}&    \hspace{-1.4mm}
\includegraphics[width=.12\textwidth]{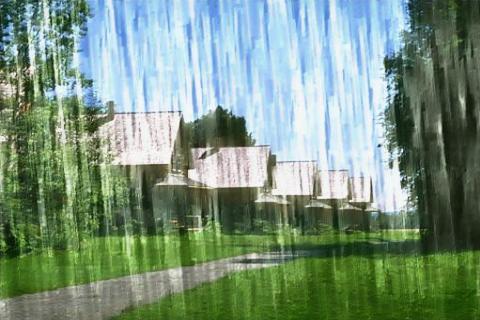}&   \hspace{-1.4mm}
\includegraphics[width=.12\textwidth]{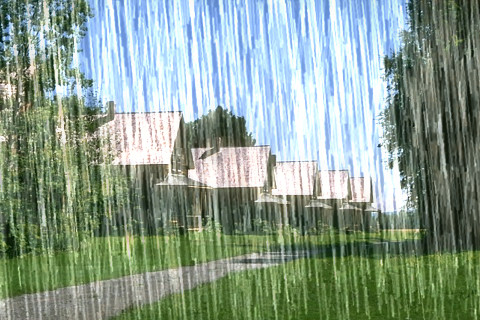}&   \hspace{-1.4mm}
\includegraphics[width=.12\textwidth]{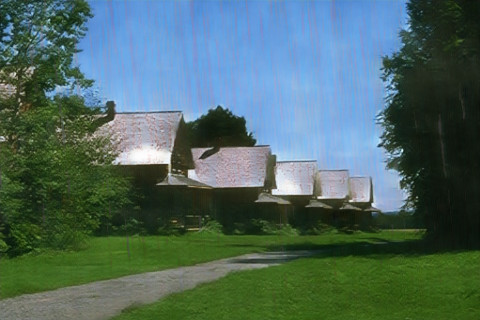}&   \hspace{-1.4mm}
\includegraphics[width=.12\textwidth]{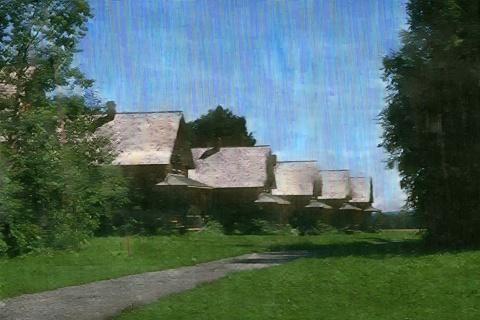}&   \hspace{-1.4mm}
\includegraphics[width=.12\textwidth]{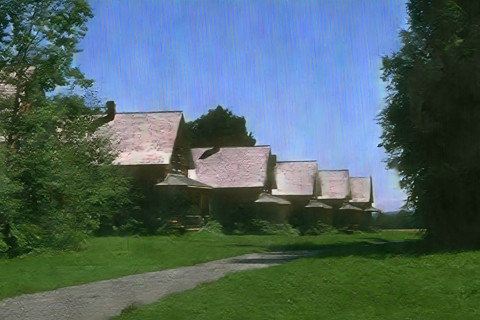}&   \hspace{-1.4mm}
\includegraphics[width=.12\textwidth]{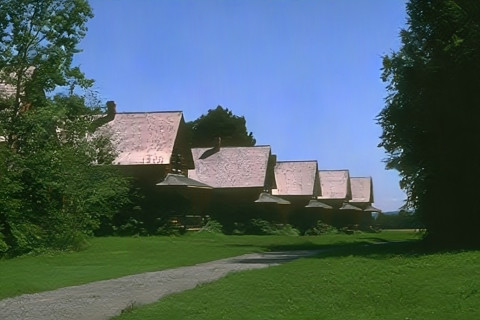}\\
\hspace{-4mm}\small~PSNR  &  \small~11.04 dB & \small~14.70 dB & \small~13.01 dB & \small~27.15 dB & \small~26.55 dB & \small~28.67 dB & \small~\textbf{30.62 dB}  \\\hspace{-4mm}
\includegraphics[width=.12\textwidth]{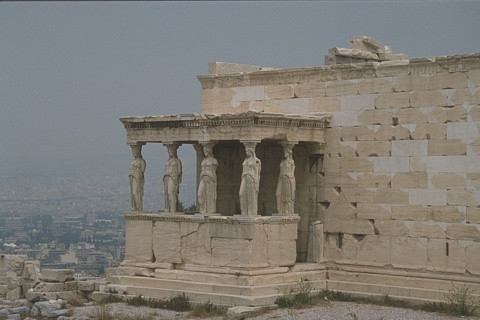}&   \hspace{-1.4mm}
\includegraphics[width=.12\textwidth]{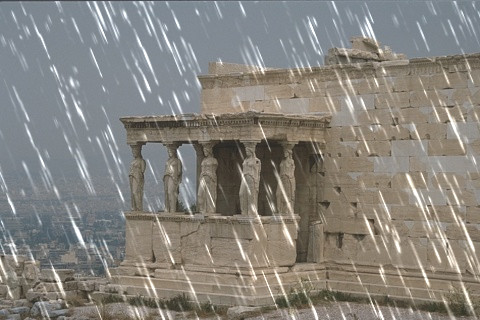}&    \hspace{-1.4mm}
\includegraphics[width=.12\textwidth]{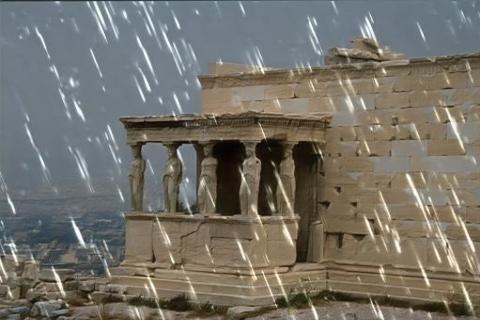}&   \hspace{-1.4mm}
\includegraphics[width=.12\textwidth]{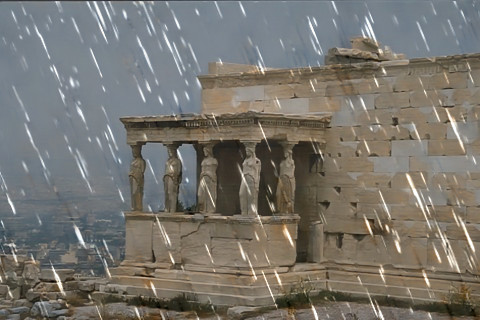}&   \hspace{-1.4mm}
\includegraphics[width=.12\textwidth]{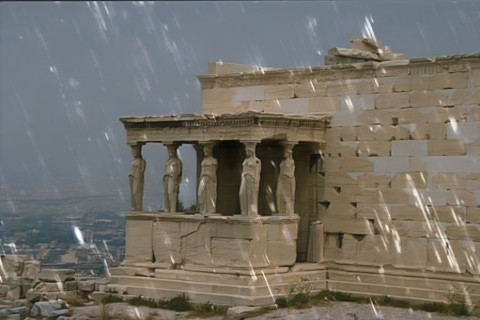}&   \hspace{-1.4mm}
\includegraphics[width=.12\textwidth]{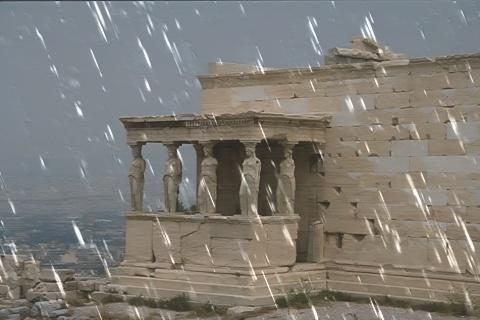}&   \hspace{-1.4mm}
\includegraphics[width=.12\textwidth]{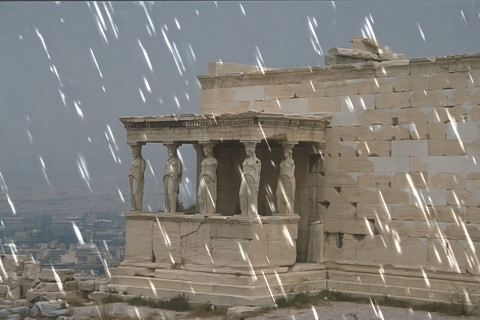}&   \hspace{-1.4mm}
\includegraphics[width=.12\textwidth]{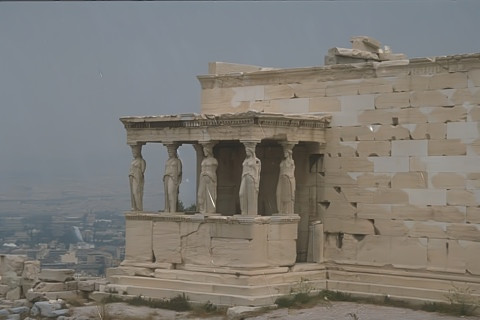}\\
\hspace{-4mm}\small~PSNR  &  \small~22.51 dB & \small~21.94 dB & \small~23.35 dB & \small~25.21 dB & \small~25.84 dB & \small~25.04 dB & \small~\textbf{38.08 dB}  \\
\hspace{-4mm} \small Reference  & \small Rainy & \small~DIDMDN~\cite{zhang2018density} & \small~SEMI~\cite{wei2019semi} & \small~UMRL~\cite{yasarla2019uncertainty}& \small~RESCAN~\cite{li2018recurrent} & \small~MSPFN~\cite{mspfn2020} & \small~\textbf{MPRNet (Ours)} \hspace{-2mm}
\end{tabular}}
\end{center}
\vspace*{-6mm}
\caption{\small Image deraining results. Our MPRNet effectively removes rain and generates images that are natural, artifact-free and visually closer to the ground-truth. 
%Additional visual examples are presented in the supplementary material.
}
\label{fig:deraining}
\vspace{-1.4em}
\end{figure*}

\subsection{Implementation Details}
Our MPRNet is end-to-end trainable and requires no pre-training.
We train separate models for three different tasks. % with the following settings.
We employ $2$ CABs at each scale of the encoder-decoder, and for downsampling we use $2$$\times$$2$ max-pooling with stride $2$. 
In the last stage, we employ ORSNet that contains $3$ ORBs, each of which further uses $8$ CABs. 
Depending on the task complexity, we scale the network width by setting the number of channels to $40$ for deraining, $80$ for denoising, and $96$ for deblurring.
The networks are trained on $256$$\times$$256$  patches with a batch size of $16$ for $4$$\times$$10^5$ iterations. 
For data augmentation, horizontal and vertical flips are randomly applied. We use Adam optimizer \cite{kingma2014adam} with the initial learning rate of $2$$\times$$10^{-4}$, which is steadily decreased to $1$$\times$$10^{-6}$ using the cosine annealing strategy~\cite{loshchilov2016sgdr}. 
%

%%%%%%%%%%%%%%%%%%%%%%%%%%%%%%%%%%%%%%%%%%%%%%%%%%%%%%%%%%%%%%%%%%%%

\subsection{Image Deraining Results}
For the image deraining task, consistent with prior work~\cite{mspfn2020}, we compute image quality scores using the Y channel (in YCbCr color space). 
Table~\ref{table:deraining} shows that our method significantly advances state-of-the-art by consistently achieving better PSNR/SSIM scores on all five datasets.
Compared to the recent best algorithm MSPFN~\cite{mspfn2020}, we obtain a performance gain of $1.98$~dB (average across all datasets), indicating $20\%$ error reduction. 
The improvements on some datasets are as large as ${4}$~dB,  e.g., Rain100L~\cite{yang2017deep}.
Further, our model has ${3.7}\times$ fewer parameters than MSPFN~\cite{mspfn2020}, while being ${2.4}\times$ faster.

Figure~\ref{fig:deraining} shows visual comparisons on challenging images. 
Our MPRNet is effective in removing rain streaks of different orientations and magnitudes, and generates images that are visually pleasant and faithful to the ground-truth.
In contrast, other approaches compromise structural content (first row), introduce artifacts (second row), and do not completely remove rain streaks (third row).

%%%%%%%%%%%%%%%%%%%%%%%%%%%%%%%%%%%%%%%%%%%%%%%%%%%%%%%%%%%%%%%%%%%
\subsection{Image Deblurring Results}
We report the performance of evaluated image deblurring approaches on the synthetic GoPro~\cite{gopro2017} and HIDE~\cite{shen2019human} datasets in Table~\ref{table:gopro and hide}.  
Overall, our model performs favorably against other algorithms.
Compared to the previous best performing technique~\cite{Maitreya2020}, our method achieves $9\%$ improvement in PSNR and $21\%$ in SSIM on the GoPro~\cite{gopro2017} dataset, and a $11\%$ and $13\%$ reduction in error on the HIDE dataset~\cite{shen2019human}. 
It is worth noticing that our network is trained only on the GoPro dataset, but achieves the state-of-the-art results (+$0.98$ dB) on the HIDE dataset, thereby demonstrating its strong generalization capability. 

We evaluate our MPRNet on the real-world images of a recent RealBlur~\cite{rim_2020_realblur} dataset under two experimental settings:~1). apply the GoPro trained model directly on RealBlur (to test generalization to real images), and 2). train and test on RealBlur data. 
Table~\ref{table:realblur} shows the experimental results. 
For setting $1$, our MPRNet obtains performance gains of $0.29$ dB on the RealBlur-R subset and $0.28$ dB on the RealBlur-J subset over the DMPHN algorithm~\cite{dmphn2019}. 
A similar trend is observed for setting $2$, where our gains over SRN~\cite{tao2018scale} are $0.66$ dB and $0.38$ dB on RealBlur-R and RealBlur-J, respectively. 

\begin{figure*}[!t]
\begin{center}
\scalebox{0.97}{
\begin{tabular}[b]{c@{ } c@{ }  c@{ } c@{ } c@{ } c@{ }	}
\hspace{-4mm}
    \multirow{4}{*}{\includegraphics[trim={ 90 0 90 0
 },clip,width=.326\textwidth,valign=t]{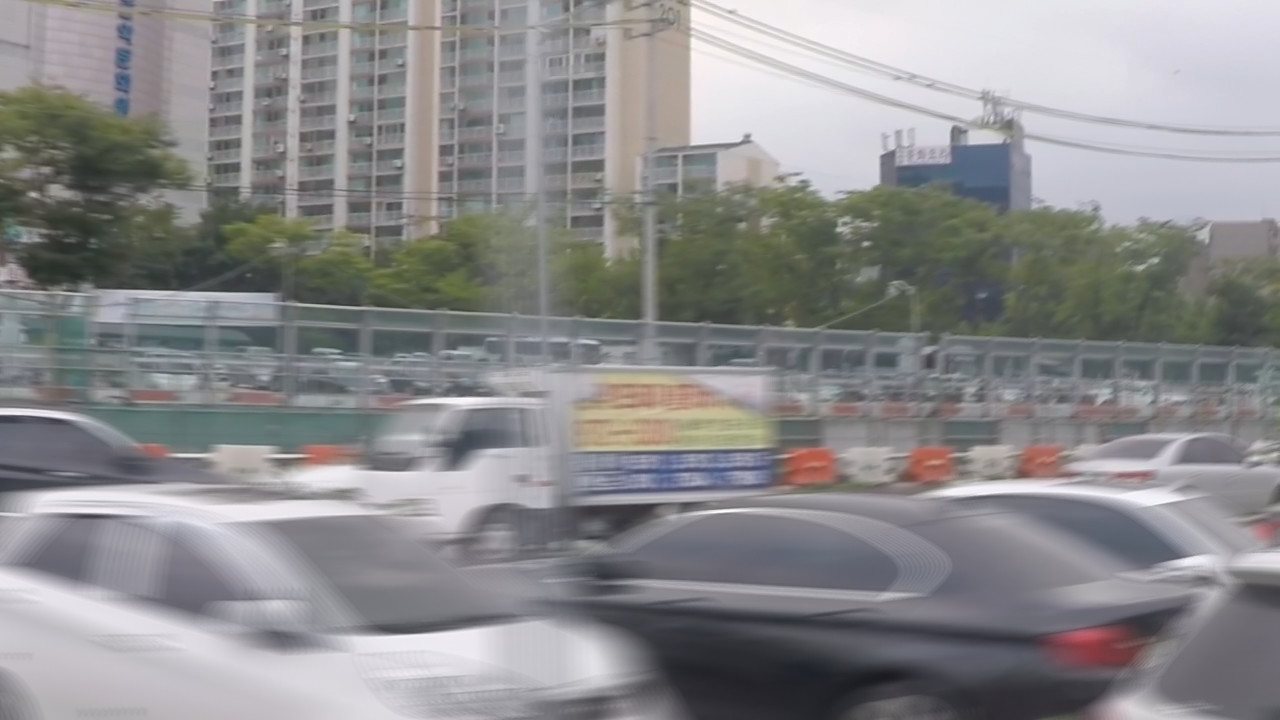}} &   
    \includegraphics[trim={ 566 220 509 370
 },clip,width=.13\textwidth,valign=t]{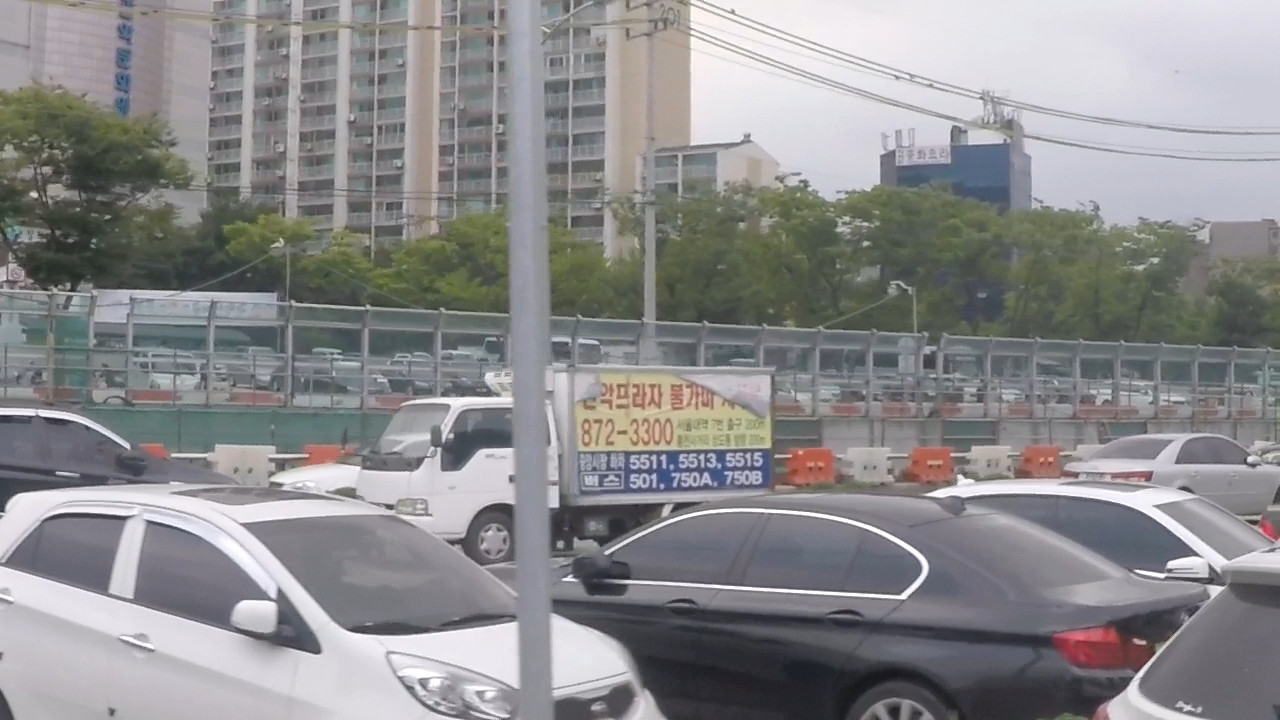}&
  	\includegraphics[trim={566 220 509 370 },clip,width=.13\textwidth,valign=t]{Images/Deblurring/img1/input.jpg}&   
    \includegraphics[trim={566 220 509 370 },clip,width=.13\textwidth,valign=t]{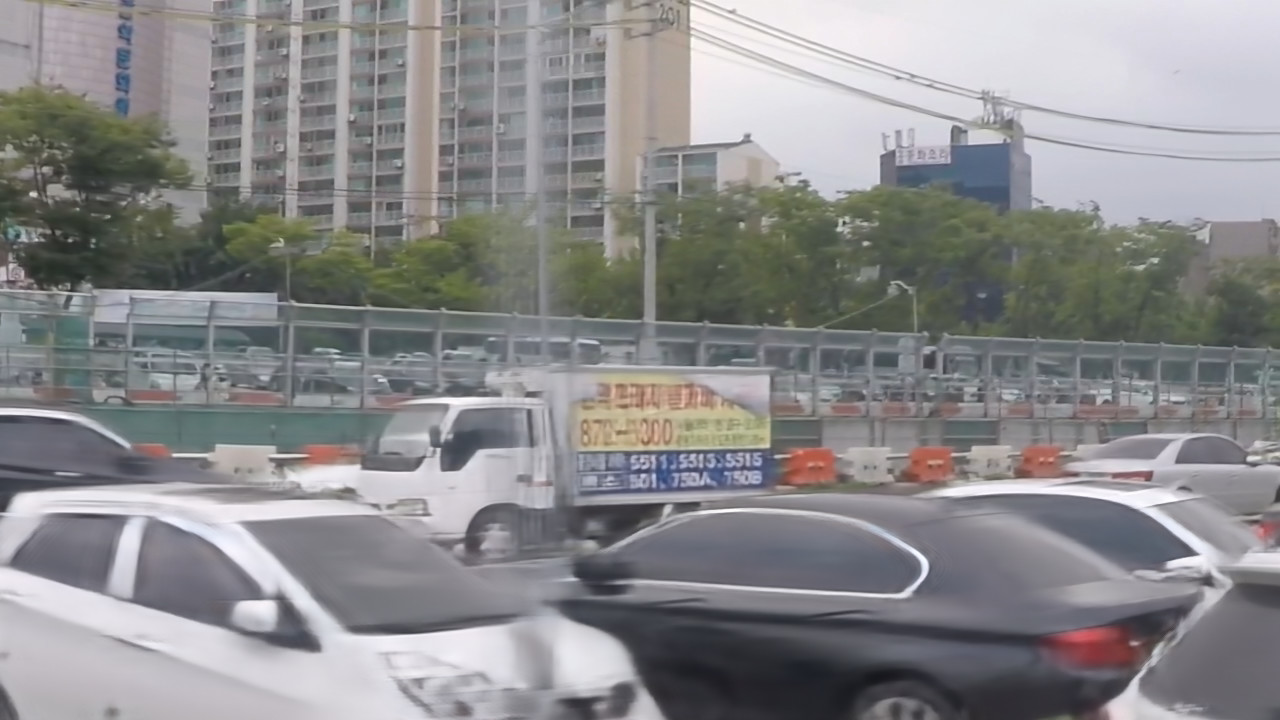}&
      	\includegraphics[trim={566 220 509 370 },clip,width=.13\textwidth,valign=t]{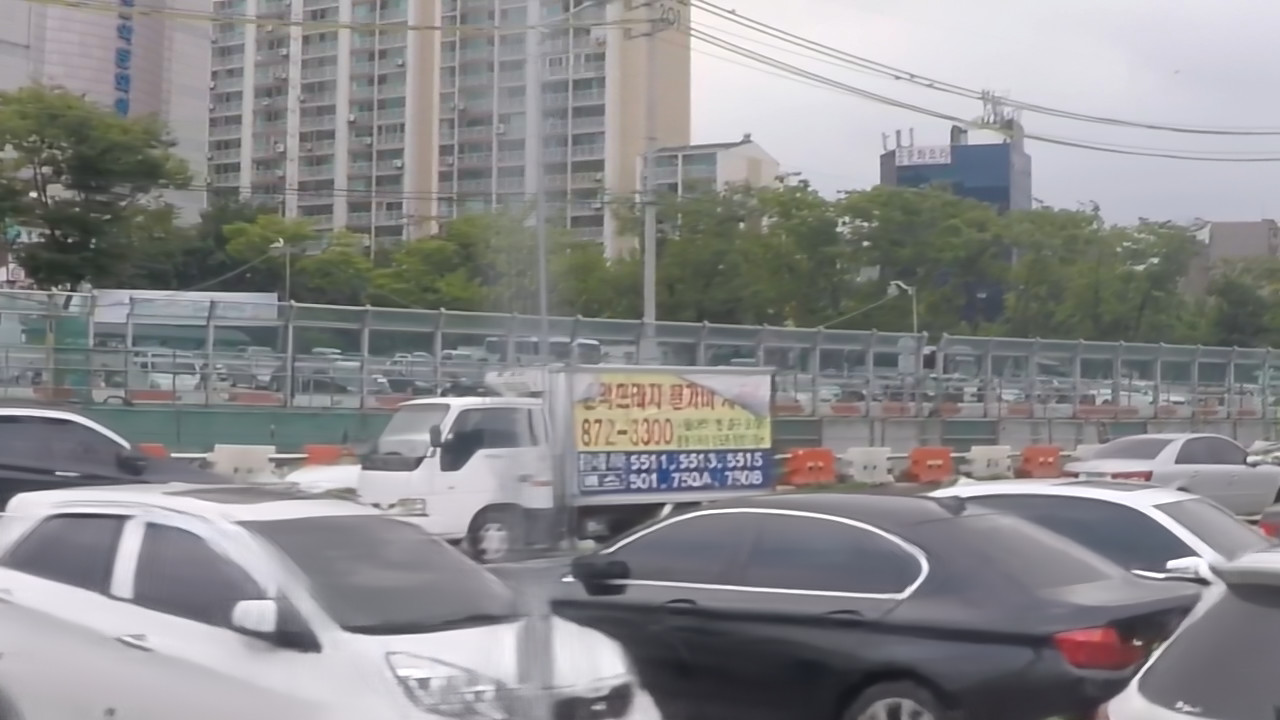}&
      \includegraphics[trim={566 220 509 370 },clip,width=.13\textwidth,valign=t]{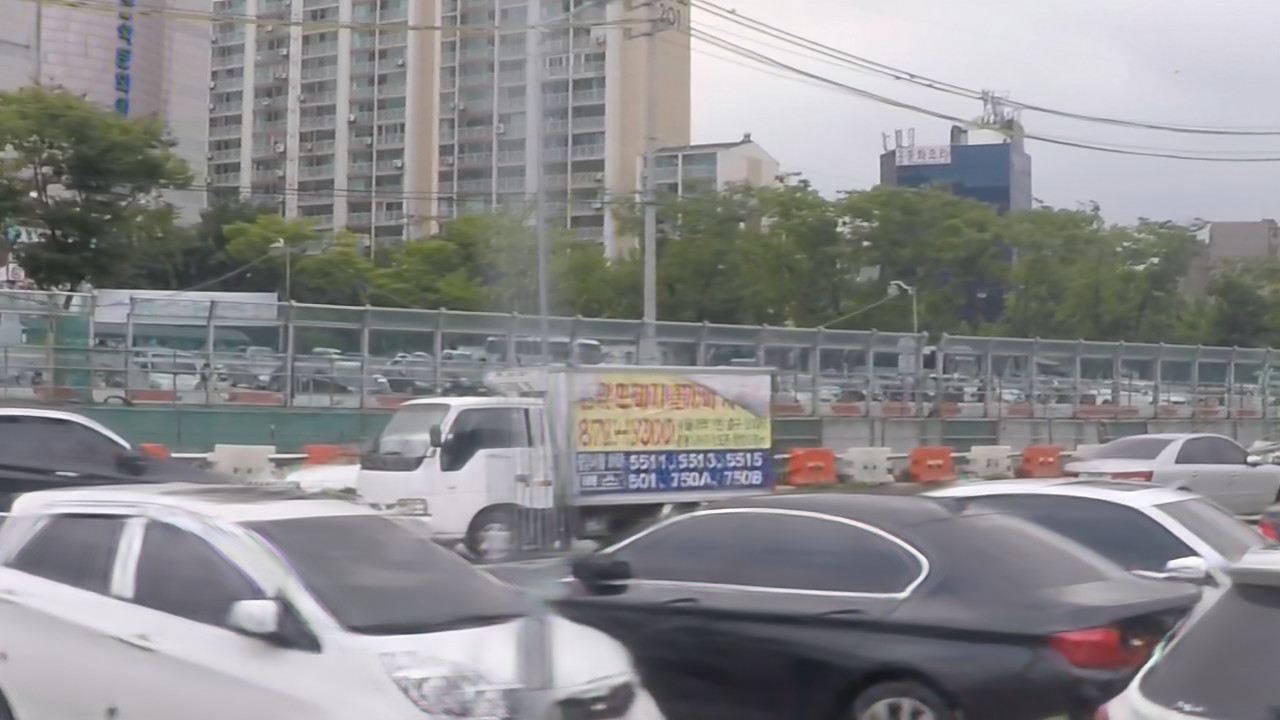}
  
\\
    &  \small~PSNR &\small~24.19 dB & \small~27.42 dB & \small~28.68 dB & \small~27.78 dB    \\
    
    & \small~Reference & \small~Blurry  & \small~SRN~\cite{tao2018scale}  & \footnotesize~DeblurGANv2~\cite{deblurganv2} & \small~Gao \etal~\cite{gao2019dynamic} \\

    &
    \includegraphics[trim={566 220 509 370 },clip,width=.13\textwidth,valign=t]{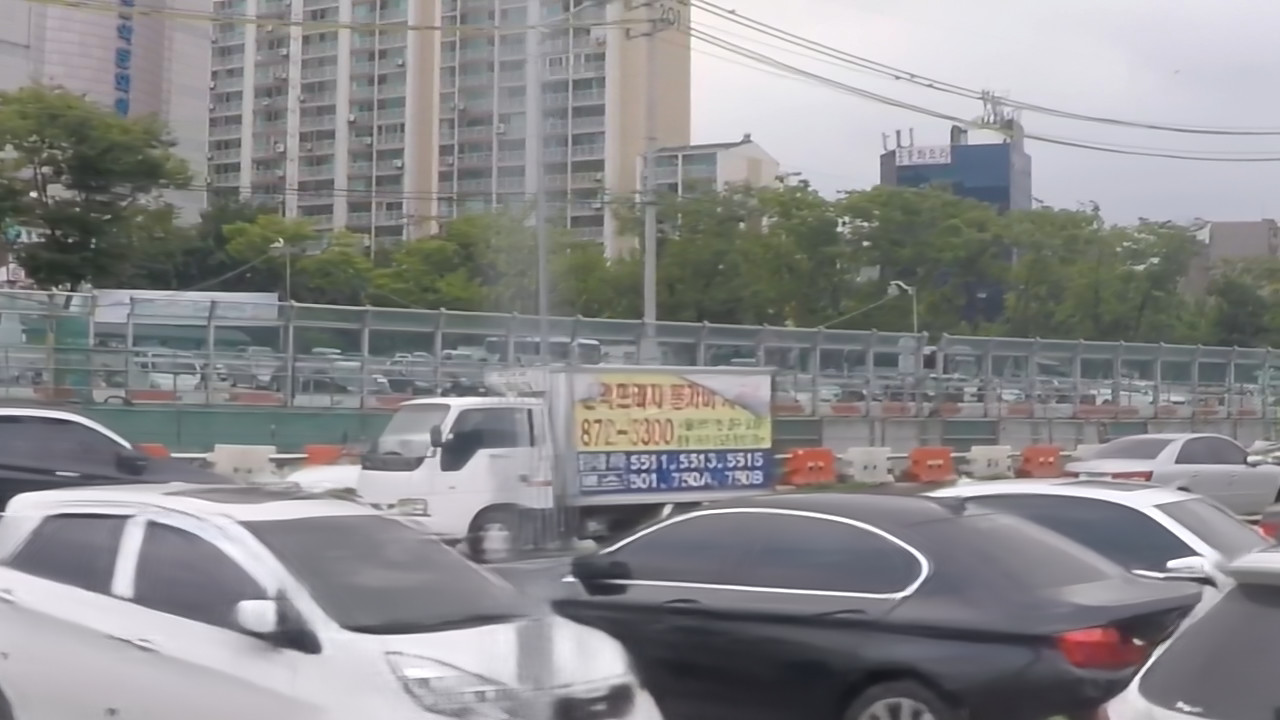}&
    \includegraphics[trim={566 220 509 370 },clip,width=.13\textwidth,valign=t]{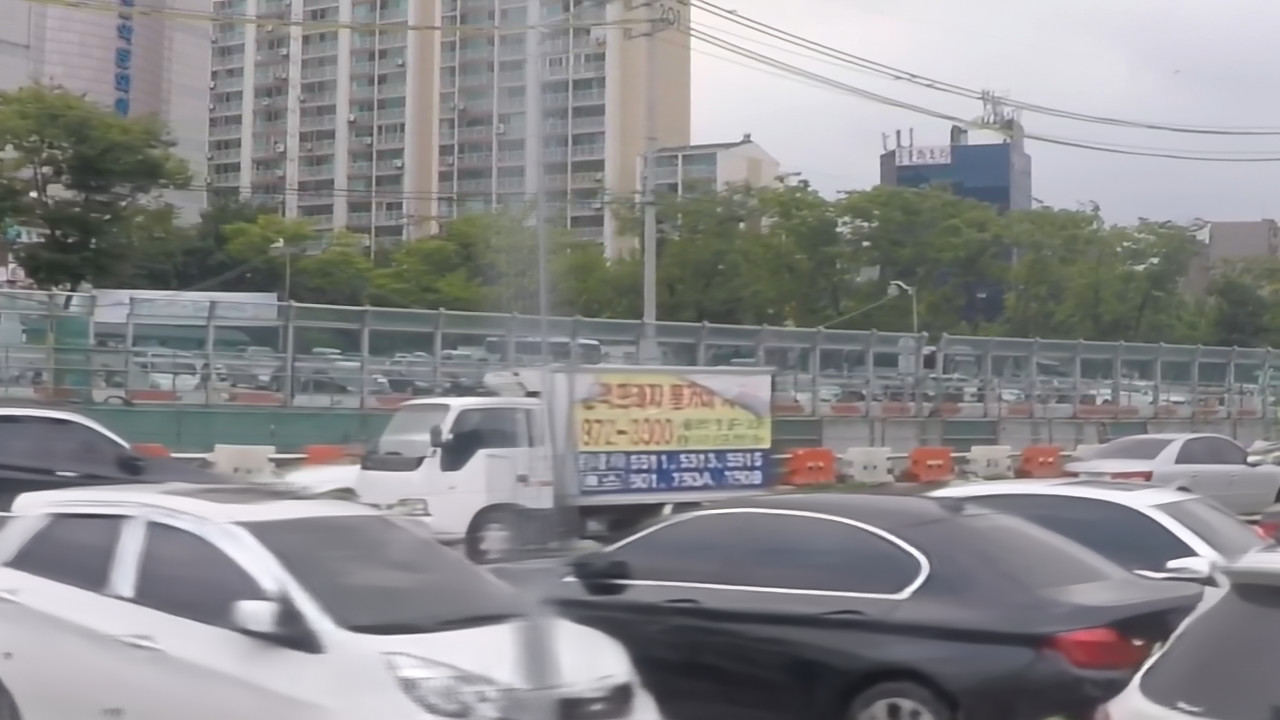}&
    \includegraphics[trim={566 220 509 370 },clip,width=.13\textwidth,valign=t]{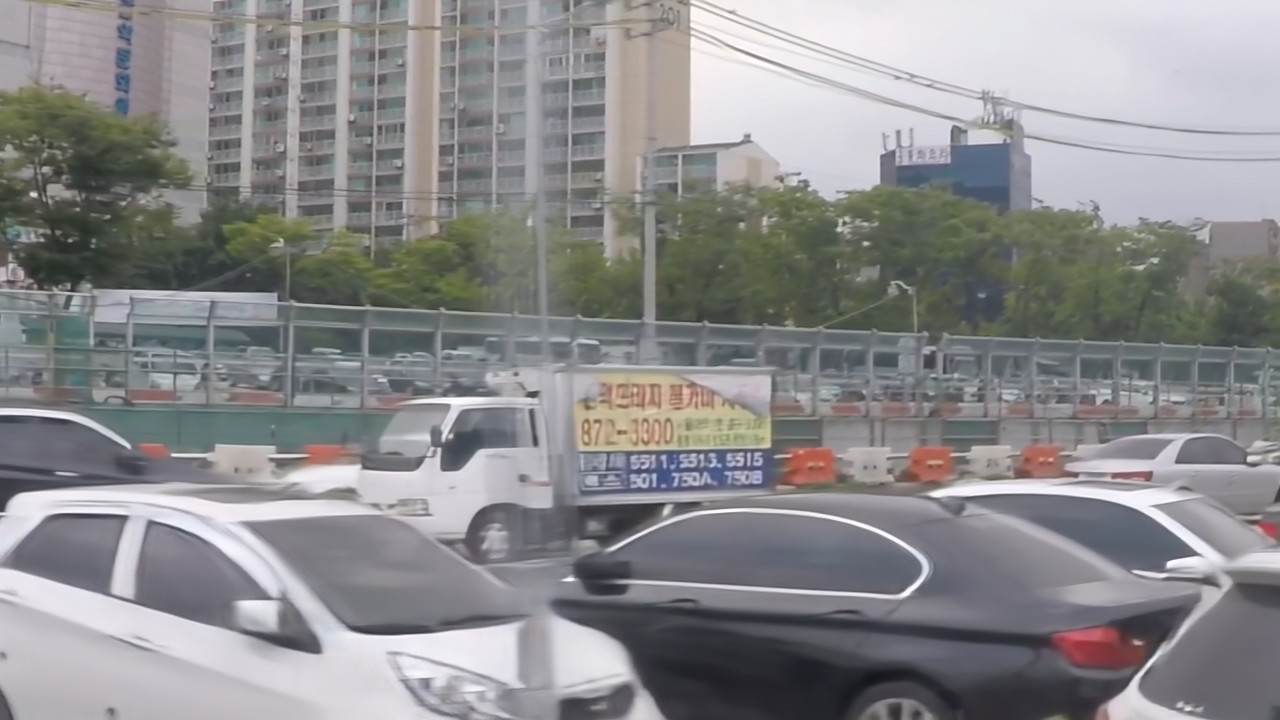}&  
     \includegraphics[trim={566 220 509 370 },clip,width=.13\textwidth,valign=t]{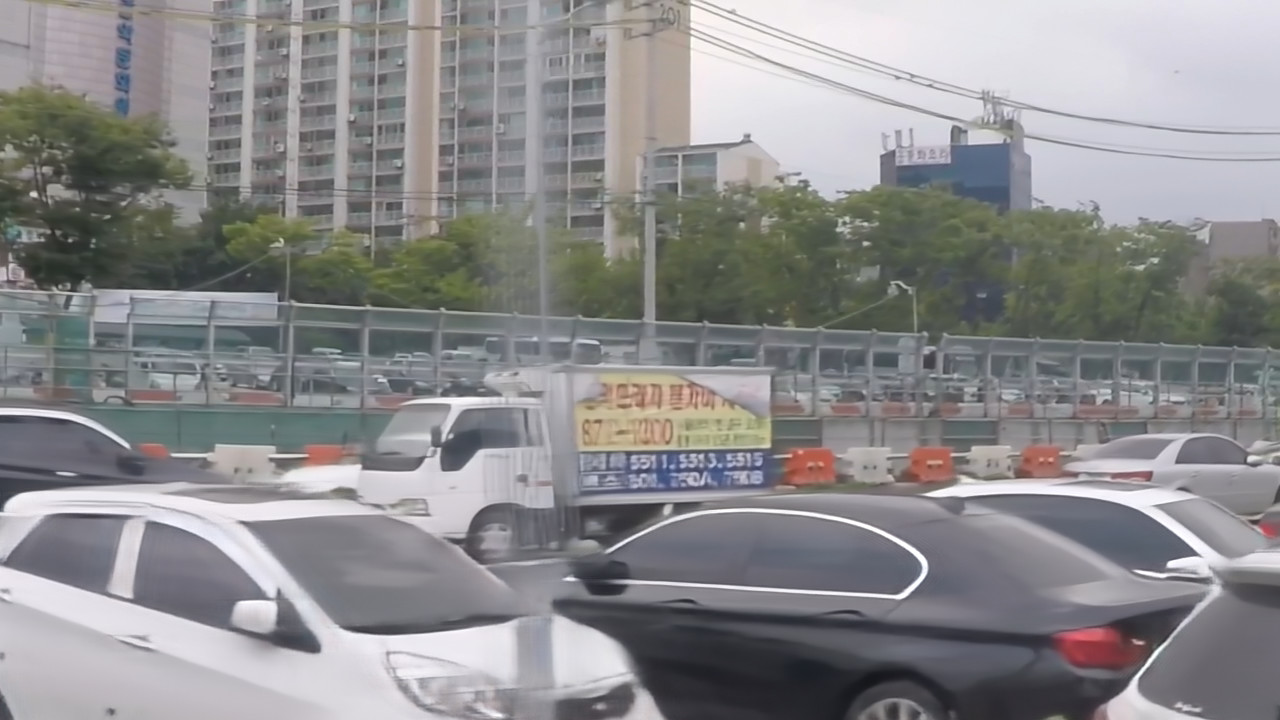}&
     \includegraphics[trim={566 220 509 370 },clip,width=.13\textwidth,valign=t]{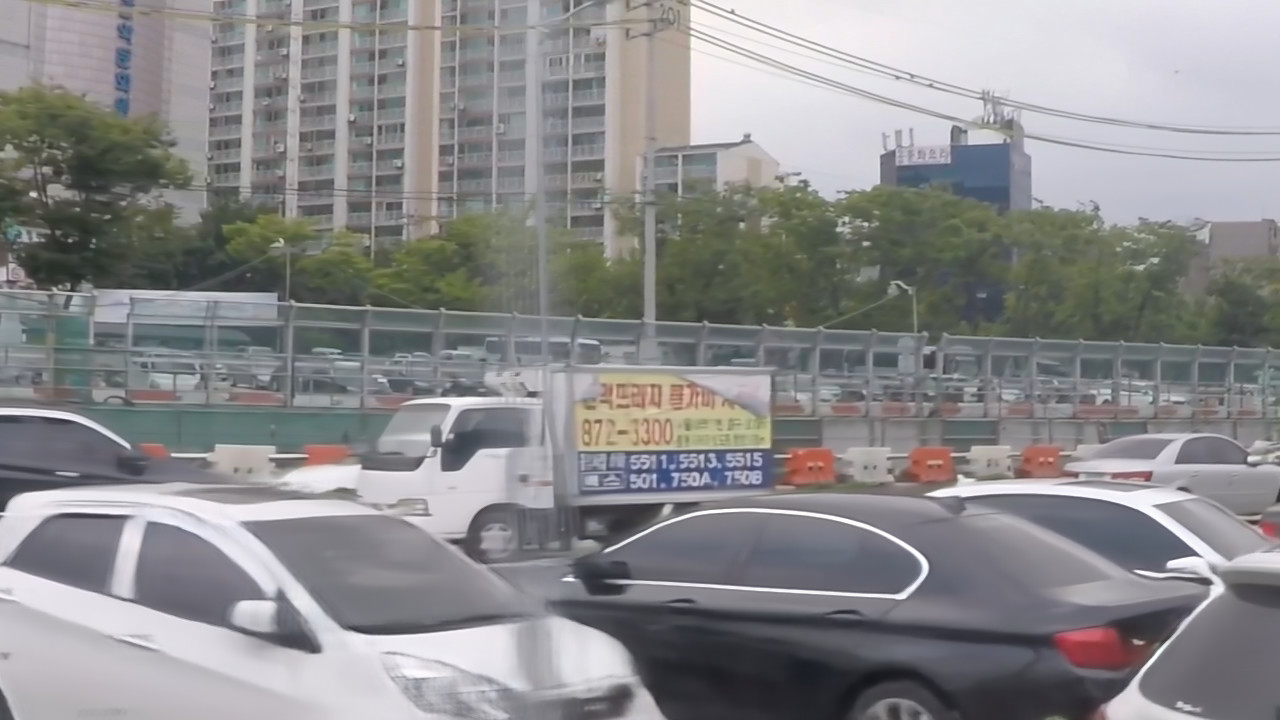}\\

     \small~24.19 dB& \small~28.60 dB& \small~28.46 dB & \small~28.54 dB 
     & \small~28.80 dB & \small~\textbf{29.16 dB}\\
           \small~Blurry Image  & \small~DBGAN~\cite{zhang2020dbgan}& \small~MTRNN~\cite{mtrnn2020} & \small~DMPHN~\cite{dmphn2019} & \small~Suin \etal~\cite{Maitreya2020}   & \small~\textbf{MPRNet (Ours)}
\vspace{3mm}
\\
\hspace{-4mm}
    \multirow{4}{*}{\includegraphics[trim={ 90 0 90 100
 },clip,width=.326\textwidth,valign=t]{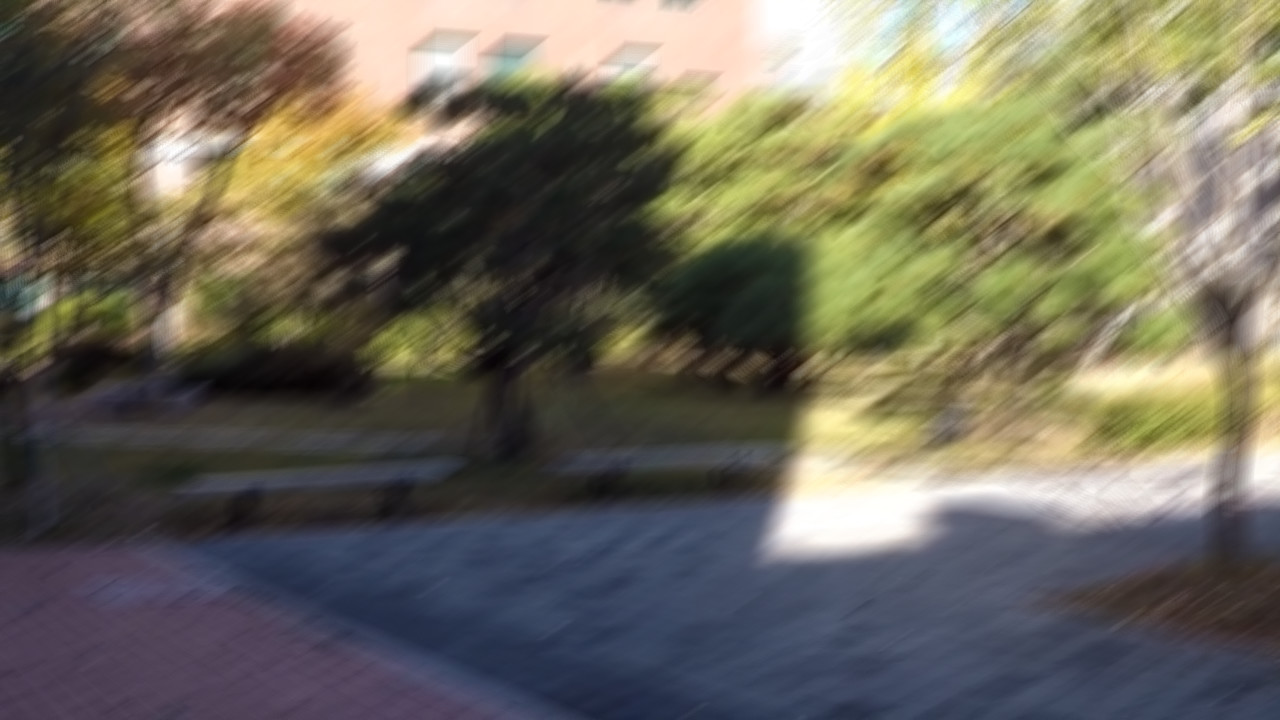}} &   
    \includegraphics[trim={ 12 60 1046 544
 },clip,width=.13\textwidth,valign=t]{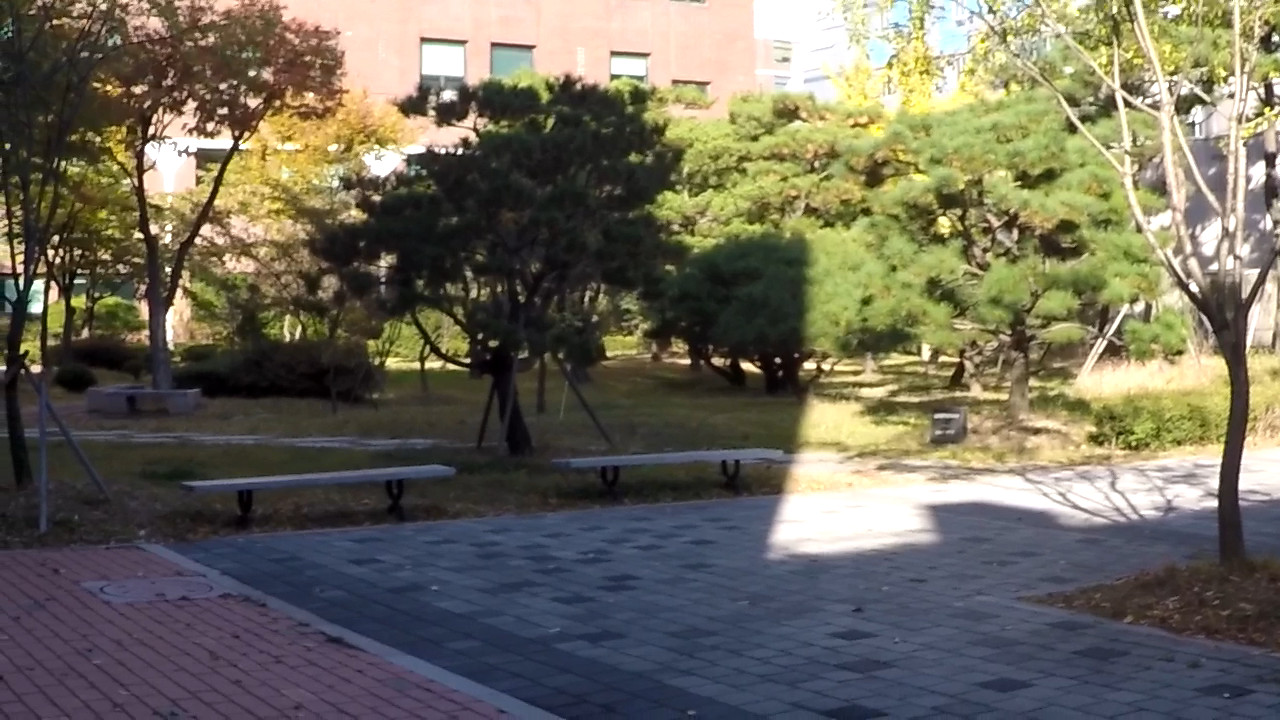}&
  	\includegraphics[trim={12 60 1046 544 },clip,width=.13\textwidth,valign=t]{Images/Deblurring/img2/input.jpg}&   
    \includegraphics[trim={12 60 1046 544 },clip,width=.13\textwidth,valign=t]{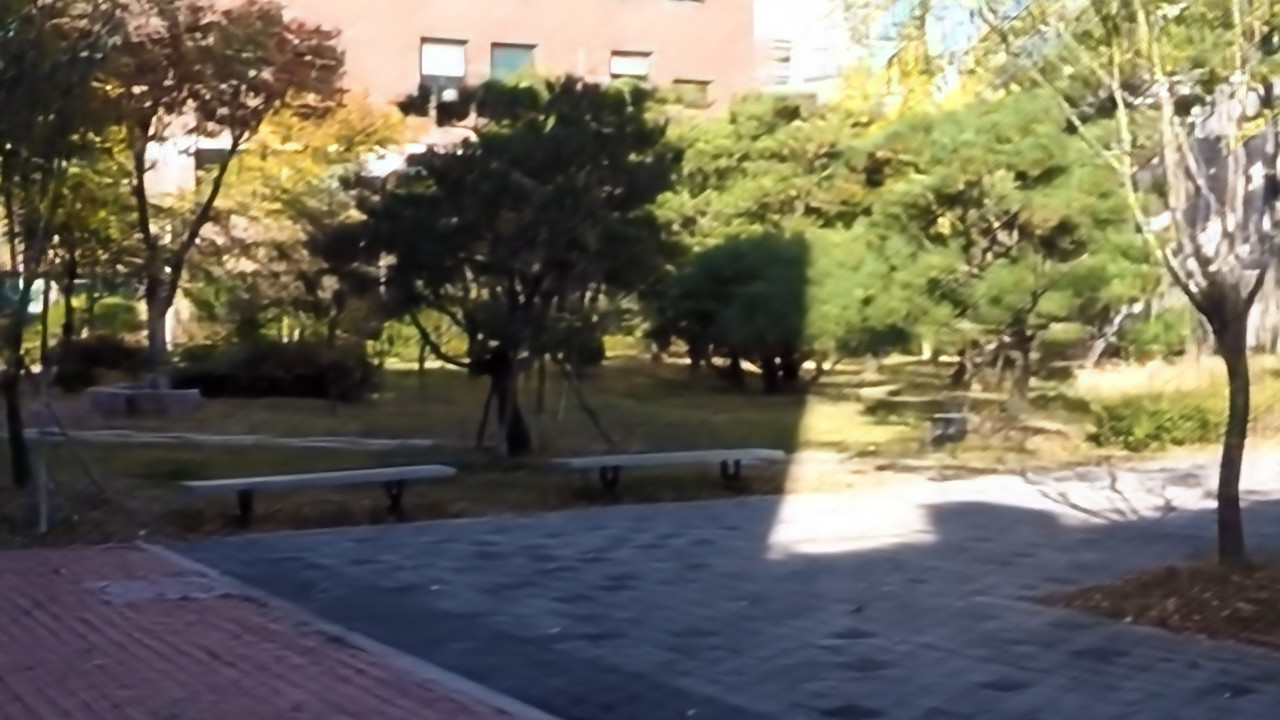}&
      	\includegraphics[trim={12 60 1046 544 },clip,width=.13\textwidth,valign=t]{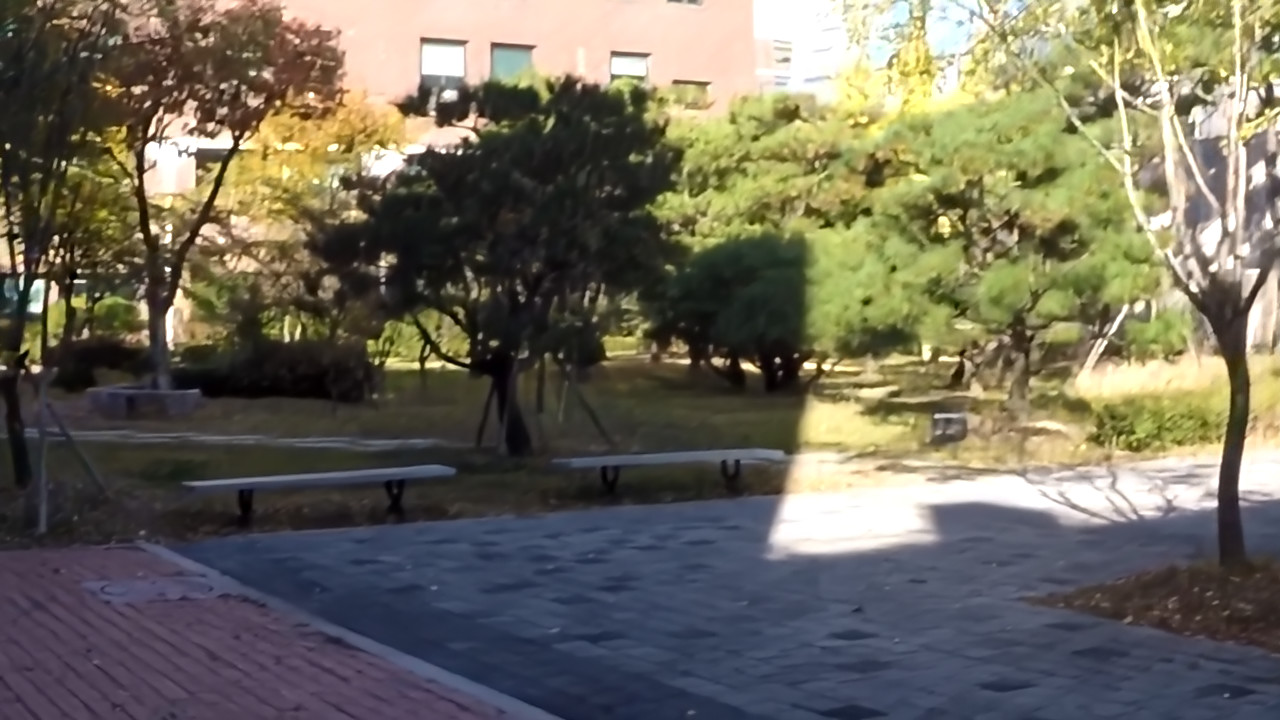}&
      \includegraphics[trim={12 60 1046 544 },clip,width=.13\textwidth,valign=t]{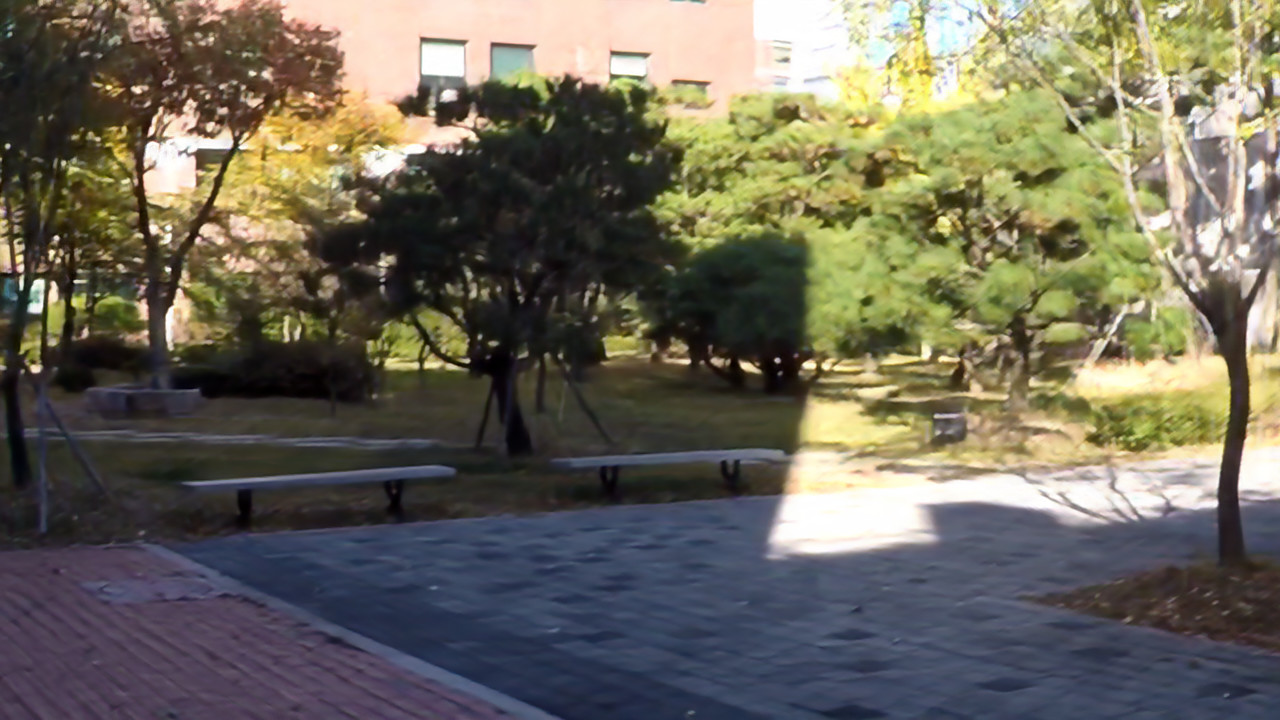}
  
\\
    &  \small~PSNR &\small~19.59 dB  & \small~24.91 dB & \small~26.30 dB & \small~26.20 dB   \\
    & \small~Reference & \small~Blurry  & \small~SRN~\cite{tao2018scale}  & \footnotesize~DeblurGANv2~\cite{deblurganv2} & \small~Gao \etal~\cite{gao2019dynamic} \\

    &
    \includegraphics[trim={12 60 1046 544 },clip,width=.13\textwidth,valign=t]{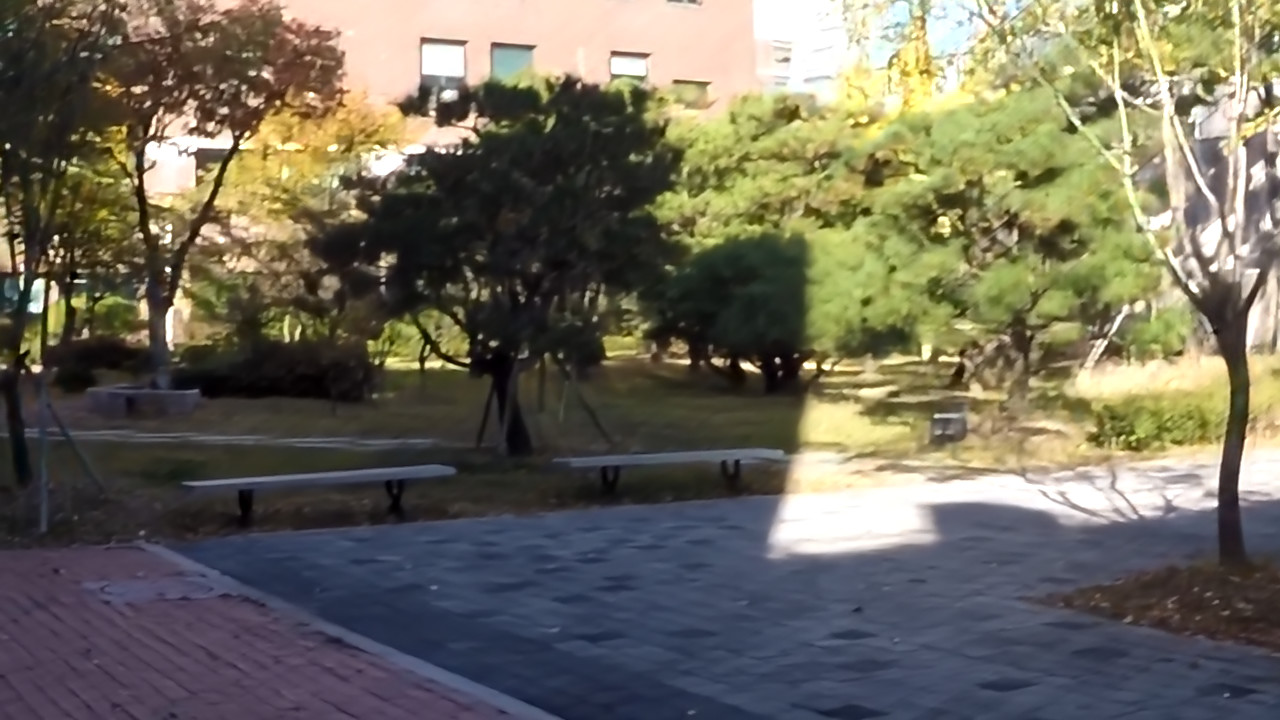}&
    \includegraphics[trim={12 60 1046 544 },clip,width=.13\textwidth,valign=t]{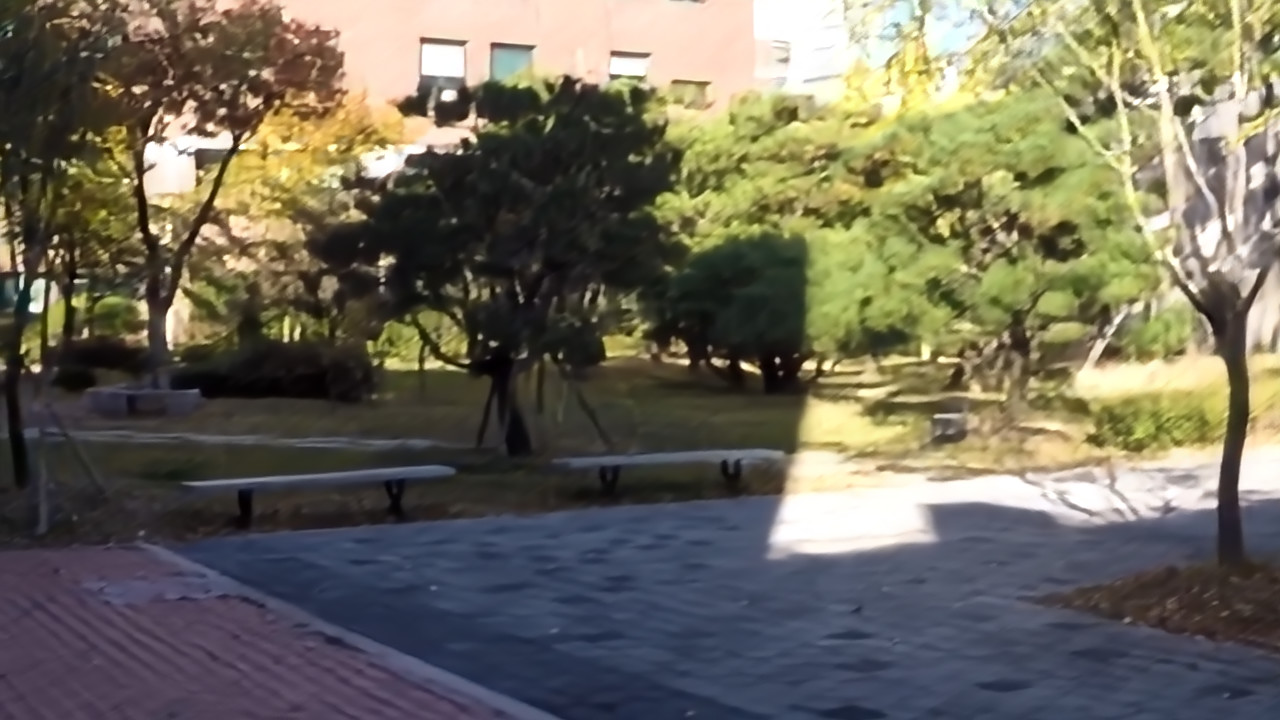}&
    \includegraphics[trim={12 60 1046 544 },clip,width=.13\textwidth,valign=t]{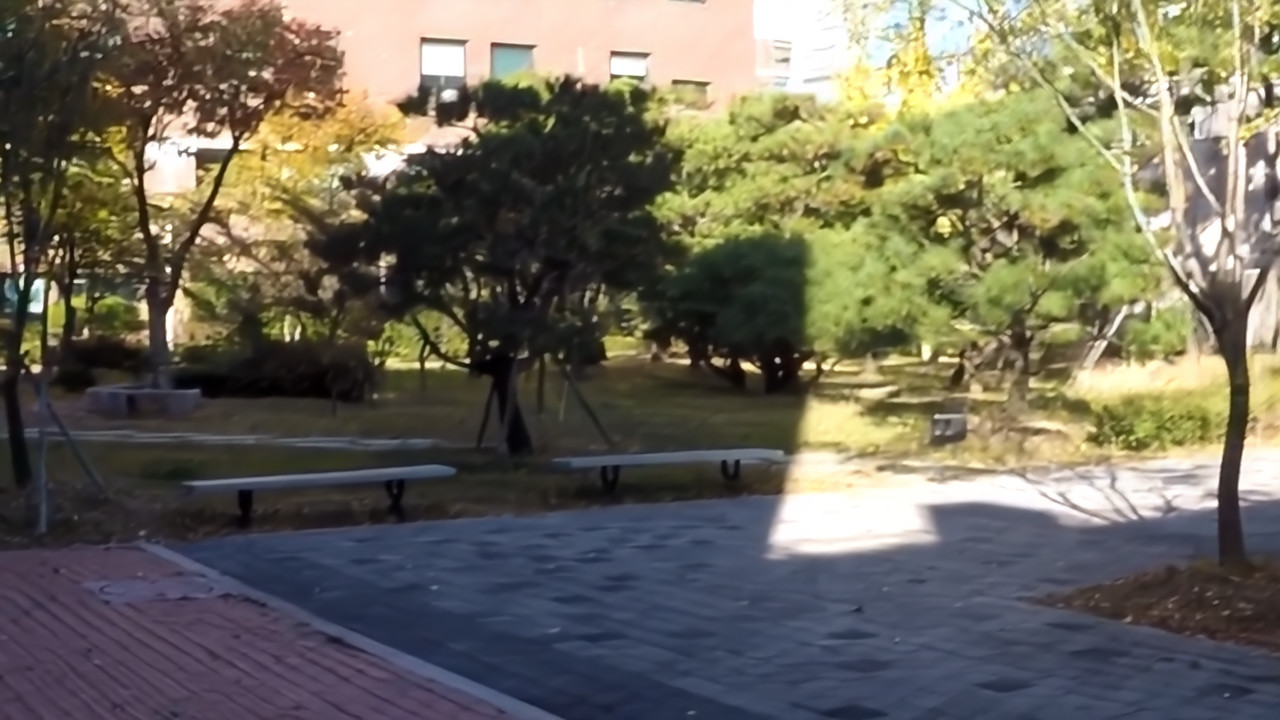}&  
     \includegraphics[trim={12 60 1046 544 },clip,width=.13\textwidth,valign=t]{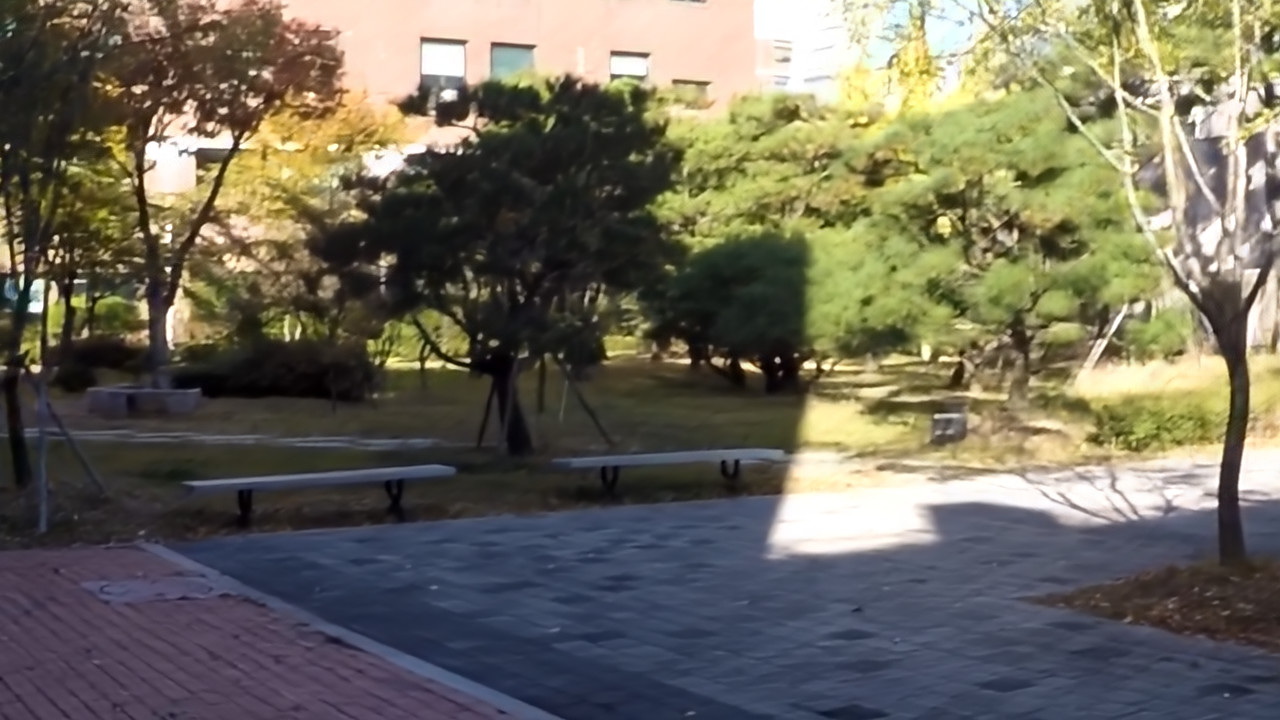}&
     \includegraphics[trim={12 60 1046 544 },clip,width=.13\textwidth,valign=t]{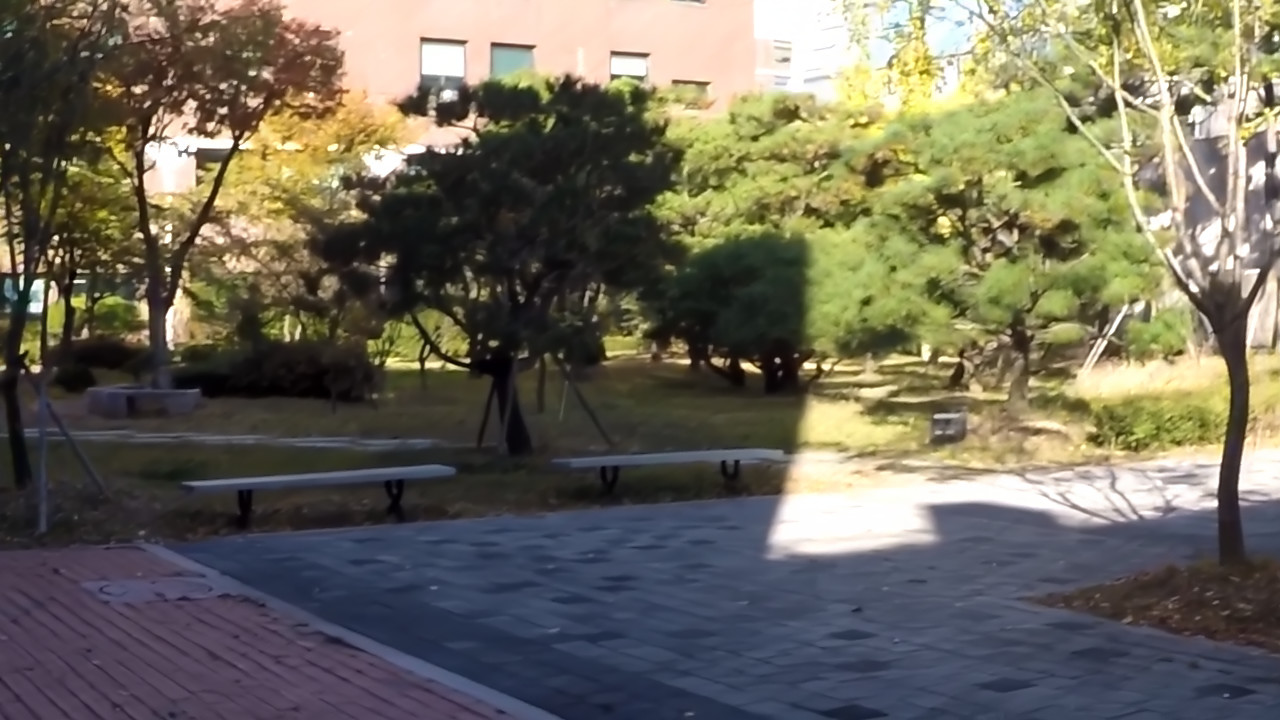}\\

     \small~19.59 dB& \small~26.15 dB& \small~25.02 dB & \small~26.66 dB 
     & \small~26.95 dB & \small~\textbf{28.68 dB}\\
           \small~Blurry Image  & \small~DBGAN~\cite{zhang2020dbgan}& \small~MTRNN~\cite{mtrnn2020} & \small~DMPHN~\cite{dmphn2019} & \small~Suin \etal~\cite{Maitreya2020}   & \small~\textbf{MPRNet (Ours)}
\\
\end{tabular}}
\end{center}
\vspace{-6mm}
\caption{ \small  Visual comparisons for image deblurring on the GoPro datatset~\cite{gopro2017}. Compared to the state-of-the-art methods, our MPRNet restores more sharper and perceptually-faithful images. %The full-resolution versions as well as more deblurring results (on the HIDE~\cite{shen2019human} and RealBlur~\cite{rim_2020_realblur} datasets) can be found in the supplementary material.  
}
\label{fig:deblurring}
\vspace{-1.6em}
\end{figure*}

Figure~\ref{fig:deblurring} shows some deblurred images by the evaluated approaches.
Overall, the images restored by our model are sharper and closer to the ground-truth than those by others.

\begin{table}[t]
\begin{center}
\caption{\small Deblurring results. Our method is trained only on the GoPro dataset~\cite{gopro2017} and directly applied to the HIDE dataset~\cite{shen2019human}. }
\label{table:gopro and hide}
\vspace{-2mm}
\setlength{\tabcolsep}{1pt}
\scalebox{0.69}{
\begin{tabular}{l c | c || c | c }
\toprule[0.15em]
 & \multicolumn{2}{c||}{GoPro~\cite{gopro2017}} & \multicolumn{2}{c}{HIDE~\cite{shen2019human}} \\
 Method & PSNR~$\textcolor{black}{\uparrow}$ & SSIM~$\textcolor{black}{\uparrow}$ & PSNR~$\textcolor{black}{\uparrow}$ & SSIM~$\textcolor{black}{\uparrow}$\\
\midrule[0.15em]
Xu \etal \cite{xu2013unnatural}     & 21.00 \colorbox{gray!20}{(73.9\%)} & 0.741 \colorbox{gray!20}{(84.2\%)} & -                               & - \\
Hyun \etal \cite{hyun2013dynamic}   & 23.64 \colorbox{gray!20}{(64.6\%)} & 0.824 \colorbox{gray!20}{(76.7\%)} & -                               & - \\
Whyte \etal \cite{whyte2012non}     & 24.60 \colorbox{gray!20}{(60.5\%)} & 0.846 \colorbox{gray!20}{(73.4\%)} & -                               & - \\
Gong \etal \cite{gong2017motion}    & 26.40 \colorbox{gray!20}{(51.4\%)} & 0.863 \colorbox{gray!20}{(70.1\%)} & -                               & - \\
DeblurGAN \cite{deblurgan}          & 28.70 \colorbox{gray!20}{(36.6\%)} & 0.858 \colorbox{gray!20}{(71.1\%)} & 24.51 \colorbox{gray!20}{(52.4\%)} &  0.871 \colorbox{gray!20}{(52.7\%)} \\
Nah \etal \cite{gopro2017}          & 29.08 \colorbox{gray!20}{(33.8\%)} & 0.914 \colorbox{gray!20}{(52.3\%)} & 25.73 \colorbox{gray!20}{(45.2\%)}                               & 0.874 \colorbox{gray!20}{(51.6\%)} \\
Zhang \etal \cite{zhang2018dynamic} & 29.19 \colorbox{gray!20}{(32.9\%)} & 0.931 \colorbox{gray!20}{(40.6\%)} & -                               & -  \\
\small{DeblurGAN-v2 \cite{deblurganv2}}    & 29.55 \colorbox{gray!20}{(30.1\%)} & 0.934 \colorbox{gray!20}{(37.9\%)} & 26.61 \colorbox{gray!20}{(39.4\%)} & 0.875 \colorbox{gray!20}{(51.2\%)} \\
SRN~\cite{tao2018scale}             & 30.26 \colorbox{gray!20}{(24.1\%)} & 0.934 \colorbox{gray!20}{(37.9\%)} & 28.36 \colorbox{gray!20}{(25.9\%)} & 0.915 \colorbox{gray!20}{(28.2\%)} \\
Shen \etal \cite{shen2019human}     & -                               & -                               & 28.89 \colorbox{gray!20}{(21.2\%)} & 0.930 \colorbox{gray!20}{(12.9\%)} \\
Gao \etal \cite{gao2019dynamic}     & 30.90 \colorbox{gray!20}{(18.3\%)} & 0.935 \colorbox{gray!20}{(36.9\%)} & 29.11\colorbox{gray!20}{(19.2\%)}  & 0.913 \colorbox{gray!20}{(29.9\%)} \\
DBGAN \cite{zhang2020dbgan}         & 31.10 \colorbox{gray!20}{(16.4\%)} & 0.942 \colorbox{gray!20}{(29.3\%)} & 28.94\colorbox{gray!20}{(20.8\%)}  & 0.915 \colorbox{gray!20}{(28.2\%)} \\
MT-RNN \cite{mtrnn2020}             & 31.15 \colorbox{gray!20}{(16.0\%)} & 0.945 \colorbox{gray!20}{(25.5\%)} & 29.15\colorbox{gray!20}{(18.8\%)}  & 0.918 \colorbox{gray!20}{(25.6\%)} \\
DMPHN \cite{dmphn2019}              & 31.20 \colorbox{gray!20}{(15.5\%)} & 0.940 \colorbox{gray!20}{(31.7\%)} & 29.09 \colorbox{gray!20}{(19.4\%)} & 0.924 \colorbox{gray!20}{(19.7\%)} \\
Suin \etal \cite{Maitreya2020}      & \underline{31.85} \colorbox{gray!20}{(8.9\%)} & \underline{0.948} \colorbox{gray!20}{(21.2\%)} & \underline{29.98} \colorbox{gray!20}{(10.7\%)} & \underline{0.930} \colorbox{gray!20}{(12.9\%)} \\
\bottomrule[0.1em]
\textbf{MPRNet (Ours)} & \textbf{32.66} \colorbox{gray!20}{(0.0\%)} & \textbf{0.959} \colorbox{gray!20}{(0.0\%)} &	\textbf{30.96} \colorbox{gray!20}{(0.0\%)}	&\textbf{0.939} \colorbox{gray!20}{(0.0\%)} \\
\bottomrule[0.1em]
\end{tabular}}
\end{center}\vspace{-1.5em}
\end{table}

\begin{table}[t]
\begin{center}
\caption{\small Deblurring comparisons on the RealBlur dataset~\cite{rim_2020_realblur} under two different settings: 1). applying our GoPro trained model directly on the RealBlur set (to evaluate generalization to real images), 2). Training and testing on RealBlur data where methods are denoted with symbol \textcolor{red}{$\ddagger$}. 
The PSNR/SSIM scores for other evaluated approaches are taken from the RealBlur benchmark~\cite{rim_2020_realblur}. }
\label{table:realblur}
\vspace{-2mm}
\setlength{\tabcolsep}{1pt}
\scalebox{0.678}{
\begin{tabular}{l c | c || c |  c }
\toprule[0.15em]
 & \multicolumn{2}{c||}{RealBlur-R} & \multicolumn{2}{c}{RealBlur-J} \\
%\cline{2-3} \cline{4-5}
 Method & PSNR~$\textcolor{black}{\uparrow}$ & SSIM~$\textcolor{black}{\uparrow}$ & PSNR~$\textcolor{black}{\uparrow}$ & SSIM~$\textcolor{black}{\uparrow}$\\
\midrule[0.15em]
Hu \etal \cite{hu2014deblurring}     &  33.67 \colorbox{gray!20}{(23.4\%)}  &  0.916 \colorbox{gray!20}{(42.9\%)}  &  26.41 \colorbox{gray!20}{(23.2\%)}  &  0.803 \colorbox{gray!20}{(35.5\%)}  \\
Nah \etal \cite{gopro2017}           &  32.51 \colorbox{gray!20}{(33.0\%)}  &  0.841 \colorbox{gray!20}{(69.8\%)}  &  27.87 \colorbox{gray!20}{(9.1\%)}  &  0.827 \colorbox{gray!20}{(26.6\%)} \\
DeblurGAN \cite{deblurgan}           &  33.79 \colorbox{gray!20}{(22.4\%)}  &  0.903 \colorbox{gray!20}{(50.5\%)}  &  27.97 \colorbox{gray!20}{(8.1\%)}  &  0.834 \colorbox{gray!20}{(23.5\%)} \\
Pan \etal \cite{pan2016blind}        &  34.01 \colorbox{gray!20}{(20.4\%)}  &  0.916 \colorbox{gray!20}{(42.9\%)}  &  27.22 \colorbox{gray!20}{(15.7\%)}  &  0.790 \colorbox{gray!20}{(39.5\%)} \\
Xu \etal \cite{xu2013unnatural}      &  34.46 \colorbox{gray!20}{(16.2\%)}  &  0.937 \colorbox{gray!20}{(23.8\%)}  &  27.14 \colorbox{gray!20}{(16.4\%)}  &  0.830 \colorbox{gray!20}{(25.3\%)} \\
\small{DeblurGAN-v2 \cite{deblurganv2}}      &  35.26 \colorbox{gray!20}{(8.1\%)}  &  0.944 \colorbox{gray!20}{(14.3\%)}   &  \underline{28.70} \colorbox{gray!20}{(0.0\%)}  &  0.866 \colorbox{gray!20}{(5.2\%)} \\
Zhang \etal \cite{zhang2018dynamic}  &  35.48 \colorbox{gray!20}{(5.7\%)}  &  0.947 \colorbox{gray!20}{(9.4\%)}    &  27.80 \colorbox{gray!20}{(9.8\%)}  &  0.847 \colorbox{gray!20}{(17.0\%)} \\
SRN~\cite{tao2018scale}              &  35.66 \colorbox{gray!20}{(3.7\%)}  &  0.947 \colorbox{gray!20}{(9.4\%)}    &  28.56  \colorbox{gray!20}{(1.6\%)} &  \underline{0.867} \colorbox{gray!20}{(4.5\%)} \\
DMPHN \cite{dmphn2019}               &  \underline{35.70} \colorbox{gray!20}{(3.3\%)} & \underline{0.948} \colorbox{gray!20}{(7.7\%)} & 28.42 \colorbox{gray!20}{(3.2\%)}  &  0.860 \colorbox{gray!20}{(9.3\%)} \\
\textbf{MPRNet~(Ours)}                        &  \textbf{35.99}	  \colorbox{gray!20}{(0.0\%)} & \textbf{0.952} \colorbox{gray!20}{(0.0\%)}    & \textbf{28.70} \colorbox{gray!20}{(0.0\%)} & \textbf{0.873} \colorbox{gray!20}{(0.0\%)} \\
\midrule
\midrule
\textcolor{red}{$^\ddagger$}\small{DeblurGAN-v2 \cite{deblurganv2}  }    &  36.44  \colorbox{gray!20}{(28.1\%)} &  0.935 \colorbox{gray!20}{(56.9\%)} &   29.69 \colorbox{gray!20}{(21.2\%)}  &  0.870 \colorbox{gray!20}{(40.0\%)}  \\
\textcolor{red}{$^\ddagger$}SRN  \cite{tao2018scale}             &  \underline{38.65} \colorbox{gray!20}{(7.3\%)}  &  \underline{0.965} \colorbox{gray!20}{(20.0\%)}   &   \underline{31.38} \colorbox{gray!20}{(4.3\%)}  &  \underline{0.909} \colorbox{gray!20}{(14.3\%)} \\
\textcolor{red}{$^\ddagger$}\textbf{MPRNet~(Ours)}  &	\textbf{39.31} \colorbox{gray!20}{(0.0\%)}	& \textbf{0.972} \colorbox{gray!20}{(0.0\%)} & \textbf{31.76} \colorbox{gray!20}{(0.0\%)} & \textbf{0.922} \colorbox{gray!20}{(0.0\%)} \\
\bottomrule[0.1em]
\end{tabular}}
\end{center}
\vspace{-1.6em}
\end{table}

%%%%%%%%%%%%%%%%%%%%%%%%%%%%%%%%%%%%%%%%%%%%%%%%%%%%%%%%%%%%%%%%%%%
\subsection{Image Denoising Results}
In Table~\ref{table:denoising}, we report PSNR/SSIM scores of several image denoising methods on the SIDD~\cite{sidd} and DND~\cite{dnd} datasets.
Our method obtains considerable gains over the state-of-the-art approaches, \ie, $0.19$ dB over CycleISP~\cite{zamir2020cycleisp} on SIDD and $0.21$ dB over SADNet~\cite{chang2020sadnet} on DND. 
Note that the DND dataset does not contain any training images, \ie, the complete publicly released dataset is just a test set. 
Experimental results on the DND benchmark with our SIDD trained model demonstrates our model generalizes well to different image domains. 

\begin{figure*}[!t]
\begin{center}
\begin{tabular}[t]{c@{ }c@{ }c@{ }c@{ }c@{ }c@{ }c@{ }c@{ }c}\hspace{-1mm}
\includegraphics[width=.105\textwidth]{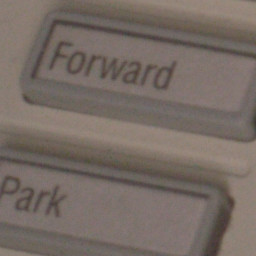}&   \hspace{-1.4mm}
\includegraphics[width=.105\textwidth]{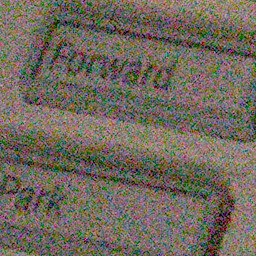}&    \hspace{-1.4mm}
\includegraphics[width=.105\textwidth]{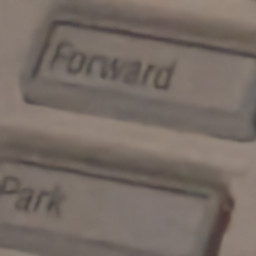}&   \hspace{-1.4mm}
\includegraphics[width=.105\textwidth]{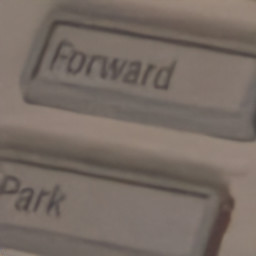}&   \hspace{-1.4mm}
\includegraphics[width=.105\textwidth]{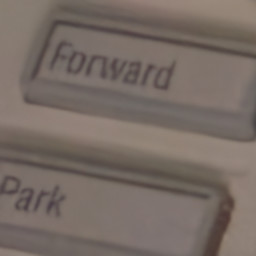}&   \hspace{-1.4mm}
\includegraphics[width=.105\textwidth]{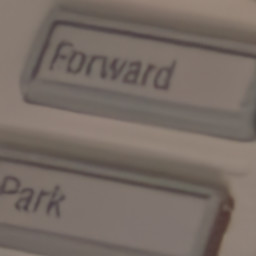}&   \hspace{-1.4mm}
\includegraphics[width=.105\textwidth]{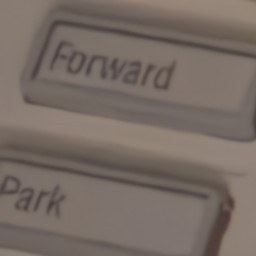}&   \hspace{-1.4mm}
\includegraphics[width=.105\textwidth]{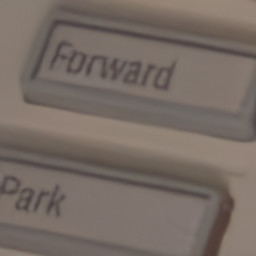}&   \hspace{-1.4mm}
\includegraphics[width=.105\textwidth]{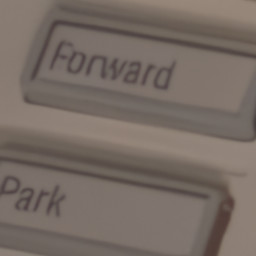}\\
\hspace{-1mm}\small~PSNR  &  \small~18.25 dB & \small~35.57 dB & \small~36.24 dB & \small~36.39 dB & \small~36.70 dB & \small~36.71 dB & \small~36.74 dB & \small~\textbf{36.98 dB}  \hspace{-4mm}\\
\includegraphics[width=.105\textwidth]{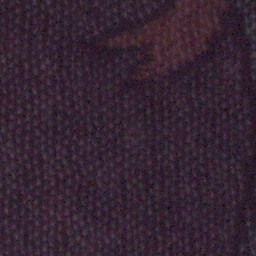}&   \hspace{-1.4mm}
\includegraphics[width=.105\textwidth]{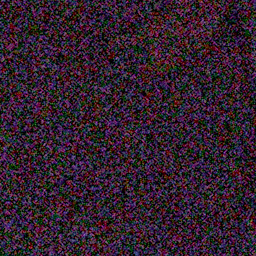}&    \hspace{-1.4mm}
\includegraphics[width=.105\textwidth]{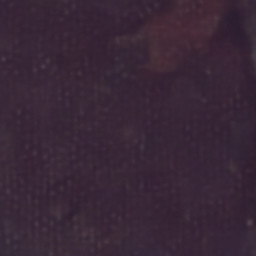}&   \hspace{-1.4mm}
\includegraphics[width=.105\textwidth]{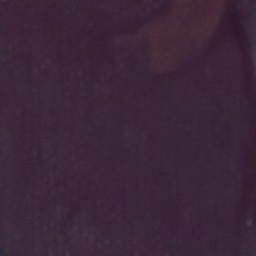}&   \hspace{-1.4mm}
\includegraphics[width=.105\textwidth]{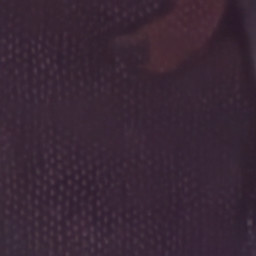}&   \hspace{-1.4mm}
\includegraphics[width=.105\textwidth]{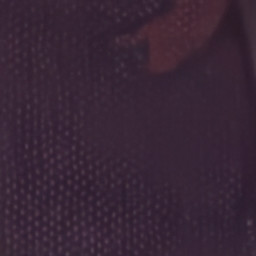}&   \hspace{-1.4mm}
\includegraphics[width=.105\textwidth]{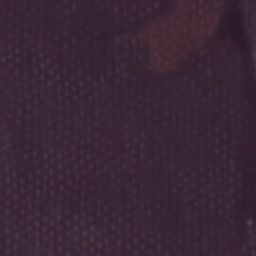}&   \hspace{-1.4mm}
\includegraphics[width=.105\textwidth]{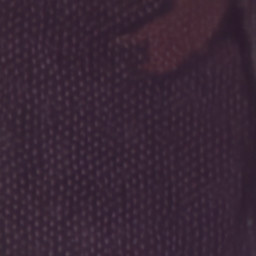}&   \hspace{-1.4mm}
\includegraphics[width=.105\textwidth]{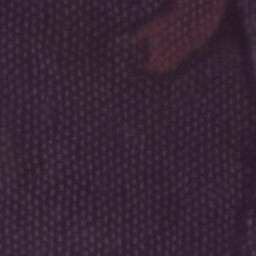}\\
\hspace{-1mm}\small~PSNR  &  \small~18.16 dB & \small~29.83 dB & \small~29.99 dB & \small~30.31 dB & \small~30.48 dB & \small~30.22 dB & \small~30.76 dB & \small~\textbf{31.17 dB}  \\
\hspace{-1mm} \small Reference  & \small Noisy & \small RIDNet~\cite{RIDNet} & \small AINDNet~\cite{kim2020aindnet} & \small VDN~\cite{VDN} &\small  SADNet~\cite{chang2020sadnet} & \small CycleISP~\cite{zamir2020cycleisp} & \small DANet~\cite{yue2020danet} &\footnotesize~\textbf{MPRNet (Ours)}  \hspace{-2mm}
\end{tabular}
\end{center}
\vspace*{-6mm}
\caption{\small Image denoising comparisons. First example is from DND~\cite{dnd} and the others from SIDD~\cite{sidd}. The proposed MPRNet better preserves fine texture and structural patterns in the denoised images.
%Additional examples are provided in the supplementary material.
}
\label{fig:denoising}
\vspace{-1.2em}
\end{figure*}

Fig.~\ref{fig:denoising} illustrates visual results. 
Our method is able to remove real noise, while preserving the structural and textural image details. 
In contrast, the images restored by other methods contain either overly smooth contents, or artifacts with splotchy textures.

\begin{table}[t]
\begin{center}
\caption{\small Denoising comparisons on SIDD~\cite{sidd} and DND~\cite{dnd} datasets. \textcolor{red}{$\ast$} denotes the methods that use additional training data. Whereas our MPRNet is only trained on the SIDD images and directly tested on DND.}
\label{table:denoising}
\vspace{-2mm}
\setlength{\tabcolsep}{1.1pt}
\scalebox{0.70}{
\begin{tabular}{l c c || c c }
\toprule[0.15em]
 & \multicolumn{2}{c||}{SIDD~\cite{sidd}} & \multicolumn{2}{c}{DND~\cite{dnd}} \\
 Method & PSNR~$\textcolor{black}{\uparrow}$ & SSIM~$\textcolor{black}{\uparrow}$ & PSNR~$\textcolor{black}{\uparrow}$ & SSIM~$\textcolor{black}{\uparrow}$\\
\midrule[0.15em]
DnCNN~\cite{DnCNN}                             & 23.66 \colorbox{gray!20}{(84.2\%)} & 0.583 \colorbox{gray!20}{(89.9\%)} & 32.43 \colorbox{gray!20}{(57.2\%)} & 0.790 \colorbox{gray!20}{(79.1\%)}\\
MLP~\cite{MLP}                                 & 24.71 \colorbox{gray!20}{(82.2\%)} & 0.641 \colorbox{gray!20}{(88.3\%)} & 34.23 \colorbox{gray!20}{(47.3\%)} & 0.833 \colorbox{gray!20}{(73.7\%)}\\
BM3D~\cite{BM3D}                               & 25.65 \colorbox{gray!20}{(80.2\%)} & 0.685 \colorbox{gray!20}{(86.7\%)} & 34.51 \colorbox{gray!20}{(45.6\%)} & 0.851 \colorbox{gray!20}{(70.5\%)}\\
CBDNet\textcolor{red}{*}~\cite{CBDNet}         & 30.78 \colorbox{gray!20}{(64.2\%)} & 0.801 \colorbox{gray!20}{(78.9\%)} & 38.06 \colorbox{gray!20}{(18.2\%)} & 0.942 \colorbox{gray!20}{(24.1\%)}\\
RIDNet\textcolor{red}{*}~\cite{RIDNet}         & 38.71 \colorbox{gray!20}{(10.9\%)} & 0.951 \colorbox{gray!20}{(14.3\%)} & 39.26 \colorbox{gray!20}{(6.0\%)} & 0.953 \colorbox{gray!20}{(6.4\%)}\\
AINDNet\textcolor{red}{*}~\cite{kim2020aindnet} & 38.95 \colorbox{gray!20}{(8.4\%)} & 0.952 \colorbox{gray!20}{(12.5\%)} & 39.37 \colorbox{gray!20}{(4.8\%)} & 0.951 \colorbox{gray!20}{(10.2\%)}\\
VDN~\cite{VDN}                                  & 39.28 \colorbox{gray!20}{(4.8\%)} & 0.956 \colorbox{gray!20}{(4.6\%)} & 39.38 \colorbox{gray!20}{(4.7\%)} & 0.952 \colorbox{gray!20}{(8.3\%)} \\
SADNet\textcolor{red}{*}~\cite{chang2020sadnet} & 39.46 \colorbox{gray!20}{(2.8\%)} & 0.957 \colorbox{gray!20}{(2.3\%)} & \underline{39.59} \colorbox{gray!20}{(2.4\%)} & 0.952 \colorbox{gray!20}{(8.3\%)} \\
DANet+\textcolor{red}{*}~\cite{yue2020danet}         & 39.47 \colorbox{gray!20}{(2.7\%)} & 0.957 \colorbox{gray!20}{(2.3\%)} & 39.58 \colorbox{gray!20}{(2.5\%)} & \underline{0.955} \colorbox{gray!20}{(2.2\%)} \\
CycleISP\textcolor{red}{*}~\cite{zamir2020cycleisp}  & \underline{39.52} \colorbox{gray!20}{(2.2\%)} & \underline{0.957} \colorbox{gray!20}{(2.3\%)} & 39.56 \colorbox{gray!20}{(2.7\%)} & \textbf{0.956}\colorbox{gray!20}{(0.0\%)}  \\
\midrule[0.1em]
 \textbf{MPRNet (Ours)} & \textbf{39.71} \colorbox{gray!20}{(0.0\%)} & \textbf{0.958} \colorbox{gray!20}{(0.0\%)} & \textbf{39.80} \colorbox{gray!20}{(0.0\%)} 	& 0.954 \colorbox{gray!20}{(4.4\%)} \\
\bottomrule[0.1em]
\end{tabular}}
\end{center}\vspace{-1.7em}
\end{table}

%%%%%%%%%%%%%%%%%%%%%%%%%%%%%%%%%%%%%%%%%%%%%%%%%%%%%%%%%%%%%%%%%%%
\subsection{Ablation Studies}

Here we present ablation experiments to analyze the contribution of each component of our model. 
Evaluation is performed on the GoPro dataset~\cite{gopro2017} with the deblurring models trained on image patches of size $128$$\times$$128$ for $10^5$ iterations, and the results are shown in   
Table~\ref{table:ablations}.

\noindent\textbf{Number of stages.} Our model yields better performance as the number of stages increases, which validates the effectiveness of our multi-stage design. 

\noindent\textbf{Choices of subnetworks.} Since each stage of our model could employ different subnetwork design, we test different options. 
We show that using the encoder-decoder in the earlier stage(s) and the ORSNet in the last stage leads to improved performance ($29.7$ dB) as compared to employing the same design for all the stages ($29.4$ dB with U-Net+U-Net, and $29.53$ dB with ORSNet+ORSNet). 

\noindent\textbf{SAM and CSFF.} We demonstrate the effectiveness of the proposed supervised attention module and cross-stage feature fusion mechanism by removing them from our final model.  
Table~\ref{table:ablations} shows a substantial drop in PSNR from $30.49$ dB to $30.07$ dB when SAM is removed, and from $30.49$ dB to $30.31$ dB when we take out CSFF. 
Removing both of these components degrades the performance by a large margin from $30.49$ dB to $29.86$ dB. 

\begin{table}[!t]
\begin{center}
\caption{Ablation study on individual components of the proposed MPRNet.}
\label{table:ablations}
\vspace{-2mm}
\setlength{\tabcolsep}{15pt}
\scalebox{0.70}{
\begin{tabular}{c  l c  c  c  }
 \toprule[0.15em]
 \#Stages   & Stage Combination  &  SAM    & CSFF  & PSNR \\
 \toprule[0.15em]
 1      &   U-Net (baseline)      & - & - & 28.94   \\
 1      &   ORSNet (baseline)       & - & - & 28.91     \\
 \midrule
 2      &   U-Net + U-Net       & \xmark & \xmark & 29.40   \\
 2      &   ORSNet + ORSNet     &  \xmark & \xmark & 29.53     \\
 2      &   U-Net + ORSNet     & \xmark & \xmark & 29.70     \\
 \midrule
 3      &   U-Nets + ORSNet     & \xmark & \xmark & 29.86     \\
 3      &   U-Nets + ORSNet     & \xmark & \cmark & 30.07     \\
 3      &   U-Nets + ORSNet     & \cmark & \xmark & 30.31     \\
 3      &   U-Nets + ORSNet     & \cmark & \cmark & \textbf{30.49}     \\
\bottomrule[0.1em]
\end{tabular}}
\end{center}\vspace{-1.7em}
\end{table}

\begin{figure*}[!t]
\begin{center}
\scalebox{0.97}{
\begin{tabular}[b]{c@{ } c@{ }  c@{ } c@{ } c@{ } c@{ }	}\hspace{-4mm}
    \multirow{4}{*}{\includegraphics[width=.314\textwidth,valign=t]{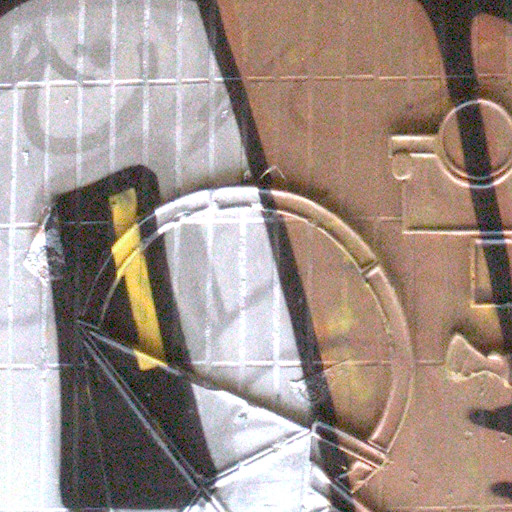}} &   
    \includegraphics[trim={10.8cm 6.5cm  3cm  7.2cm },clip,width=.13\textwidth,valign=t]{Images/Denoising/DND/noisy_26_90.jpg}&
  	\includegraphics[trim={10.8cm 6.5cm  3cm  7.2cm },clip,width=.13\textwidth,valign=t]{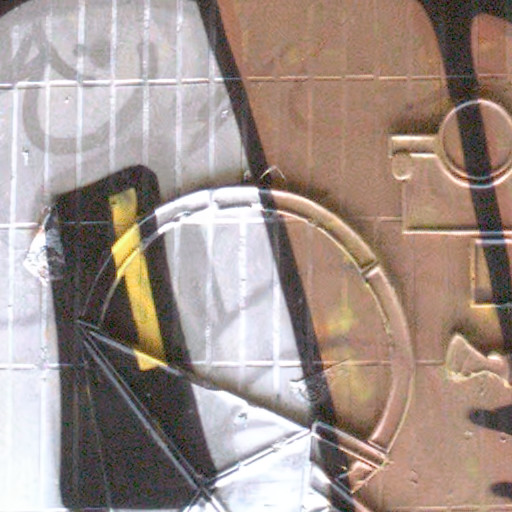}&   
    \includegraphics[trim={10.8cm 6.5cm  3cm  7.2cm },clip,width=.13\textwidth,valign=t]{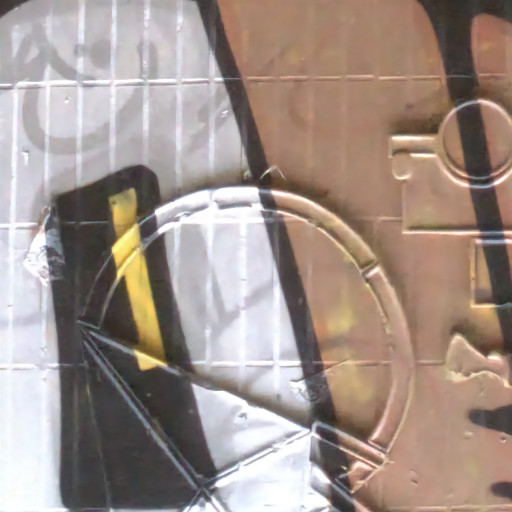}&
      	\includegraphics[trim={10.8cm 6.5cm  3cm  7.2cm },clip,width=.13\textwidth,valign=t]{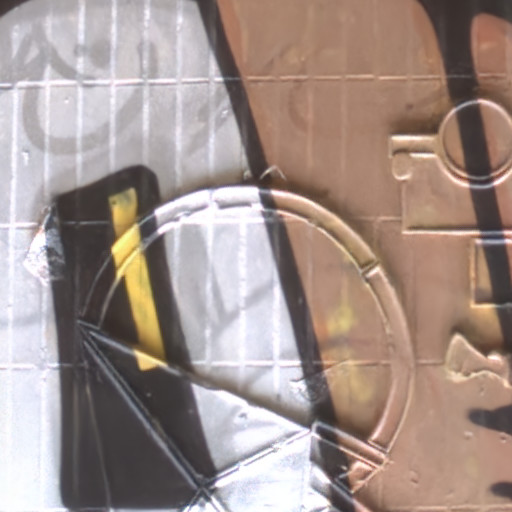}&
      \includegraphics[trim={10.8cm 6.5cm  3cm  7.2cm },clip,width=.13\textwidth,valign=t]{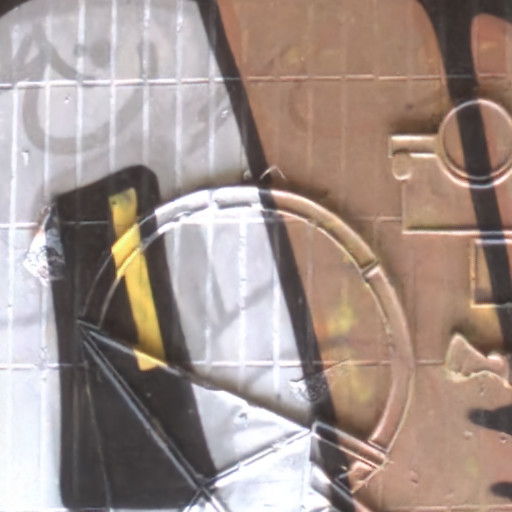}
  
\\
    &  \small~26.90 dB &\small~30.91 dB  & \small~33.62 dB & \small~33.89 dB & \small~34.09 dB   \\
    & \small~Noisy & \small~BM3D~\cite{BM3D}  & \small~CBDNet~\cite{CBDNet}  & \small~VDN~\cite{VDN} & \small~RIDNet~\cite{RIDNet} \\

    &
    \includegraphics[trim={10.8cm 6.5cm  3cm  7.2cm },clip,width=.13\textwidth,valign=t]{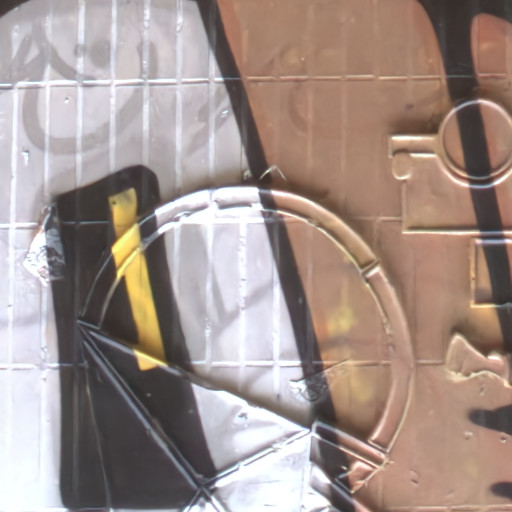}&
    \includegraphics[trim={10.8cm 6.5cm  3cm  7.2cm },clip,width=.13\textwidth,valign=t]{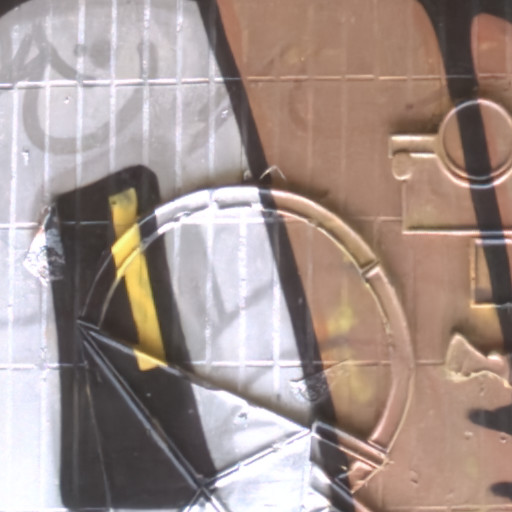}&
    \includegraphics[trim={10.8cm 6.5cm  3cm  7.2cm },clip,width=.13\textwidth,valign=t]{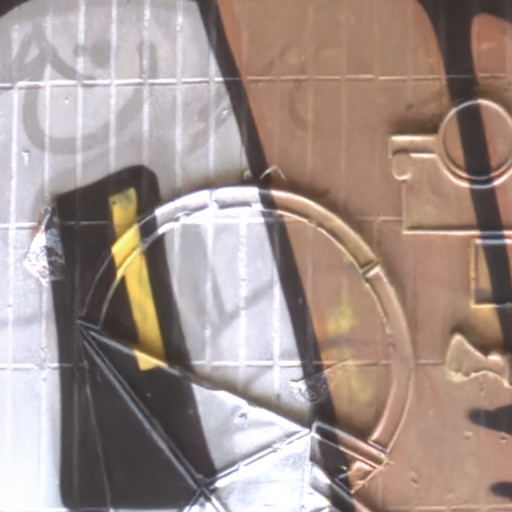}&  
     \includegraphics[trim={10.8cm 6.5cm  3cm  7.2cm },clip,width=.13\textwidth,valign=t]{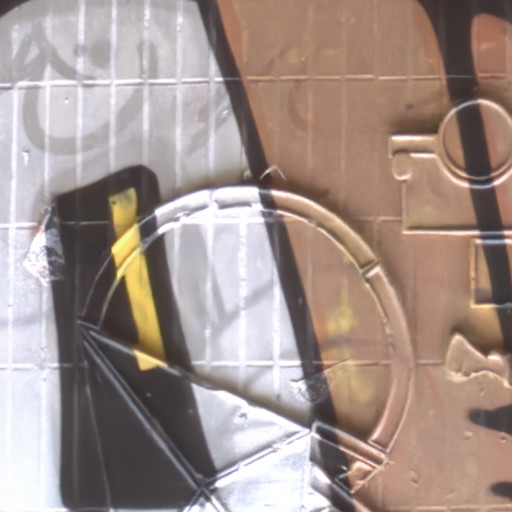}&
     \includegraphics[trim={10.8cm 6.5cm  3cm  7.2cm },clip,width=.13\textwidth,valign=t]{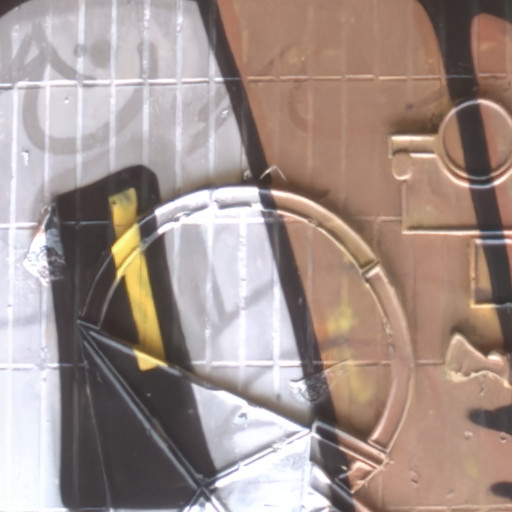}\\

     \small~26.90 dB& \small~34.32 dB& \small~34.36 dB & \small~34.36 dB 
     & \small~34.52 dB & \small~\textbf{34.91 dB}\\
           \small~Noisy Image  & \small~CycleISP~\cite{zamir2020cycleisp}& \small~AINDNet~\cite{kim2020aindnet} & \small~DANet~\cite{yue2020danet} & \small~SADNet~\cite{chang2020sadnet}   & \small~\textbf{MPRNet \footnotesize(Ours)}
\\
\end{tabular}}
\end{center}
\vspace{-6mm}
\end{figure*}

%%%%%%%%%%%%%%%%%%%%%%%%%%%%%%%%%%%%%%%%%%%%%%%%%%%%%%%%%%%%%%%%%
\section{Resource Efficient Image Restoration}
\label{sec:anystage}

CNN models generally exhibit a trade-off between accuracy and computational efficiency. 
In the pursuit of achieving higher accuracy, deeper and complex models are often developed.
Although large models tend to perform better than their smaller counterparts, the computational cost can be prohibitively high. 
As such, it is of great interest to develop resource-efficient image restoration models. 
One solution is to train the same network by adjusting its capacity every time the target system is changed.
However, it is tedious and oftentimes infeasible. 
A more desirable approach is to have a single network that can make \textbf{(a)} early predictions for compute efficient systems and \textbf{(b)} latter predictions to obtain high accuracy.  
A \emph{multi-stage} restoration model naturally offers such functionalities.  

\begin{table}[!t]
\begin{center}
\caption{\small Stage-wise deblurring performance of MPRNet on GoPro~\cite{gopro2017}. 
Runtimes are computed with the Nvidia Titan Xp GPU. }
\label{table:stagewise}
\vspace{-2mm}
\setlength{\tabcolsep}{1.7pt}
\scalebox{0.70}{
\begin{tabular}{l  cccc | c|c|c}
 \toprule[0.15em]
Method       & DeblurGAN-v2 & SRN          & DMPHN     & Suin   & \multicolumn{3}{c}{\textbf{MPRNet (ours)} }    \\
             & \cite{deblurganv2}  & \cite{tao2018scale} & \cite{dmphn2019} & \etal   \cite{Maitreya2020} & 1-stage & 2-stages & 3-stages \\
 \midrule[0.15em]
PSNR         & 29.55        & 30.10        & 31.20     & 31.85        & 30.43   & 31.81   & 32.66   \\
\#Params (M) & 60.9         & 6.8          & 21.7      & 23.0         & 5.6     & 11.3    & 20.1    \\
Time (s)     & 0.21         & 0.57          & 1.07      & 0.34         & 0.04    & 0.08    & 0.18  \\ 
\bottomrule[0.1em]
\end{tabular}}
\end{center}\vspace{-1.8em}
\end{table}

Table~\ref{table:stagewise} reports the stage-wise results of our multi-stage approach.
Our MPRNet demonstrates competitive restoration performance at each stage. 
Notably, our stage-1 model is light, fast, and yields better results than other sophisticated algorithms such as SRN~\cite{tao2018scale} and DeblurGAN-v2~\cite{deblurganv2}.
Similarly, when compared to a recent method DMPHN~\cite{dmphn2019}, our stage-2 model shows the PSNR gain of 0.51 dB while being more resource-efficient ($\sim$${2\times}$ fewer parameters and ${13\times}$ faster). 

%%%%%%%%%%%%%%%%%%%%%%%%%%%%%%%%%%%%%%%%%%%%%%%%%%%%%%%%%%%%%%%%%
\section{Conclusion}
In this work, we propose a multi-stage architecture for image restoration that progressively improves degraded inputs by injecting supervision at each stage. We develop guiding principles for our design that demand complementary feature processing in multiple stages and a flexible information exchange between them. To this end, we propose contextually-enriched and spatially accurate stages that encode a diverse set of features in unison. To ensure synergy between reciprocal stages, we propose feature fusion across stages and an attention guided output exchange from earlier stages to the later ones. Our model achieves significant performance gains on numerous benchmark datasets. In addition, our model is light-weighted in terms of model size and efficient in terms of runtime, which are of great interest for devices with limited resources. 
%
%Furthermore, our design allows an on-demand accuracy-complexity trade-off without any retraining.

\vspace{0.5em}\noindent\textbf{Acknowledgments.} M.-H. Yang is supported in part by the NSF CAREER Grant 1149783. Special thanks to Kui Jiang for providing image deraining results.

{\small
\bibliographystyle{ieee}
\bibliography{bib}

\begin{thebibliography}{100}\itemsep=-1pt

\bibitem{sidd}
Abdelrahman Abdelhamed, Stephen Lin, and Michael~S Brown.
\newblock A high-quality denoising dataset for smartphone cameras.
\newblock In {\em CVPR}, 2018.

\bibitem{ntire2019_denoising}
Abdelrahman Abdelhamed, Radu Timofte, and Michael~S Brown.
\newblock {NTIRE} 2019 challenge on real image denoising: Methods and results.
\newblock In {\em CVPRW}, 2019.

\bibitem{KSVD}
Michal Aharon, Michael Elad, and Alfred Bruckstein.
\newblock {K-SVD:} an algorithm for designing overcomplete dictionaries for
  sparse representation.
\newblock {\em Trans. Sig. Proc.}, 2006.

\bibitem{RIDNet}
Saeed Anwar and Nick Barnes.
\newblock Real image denoising with feature attention.
\newblock {\em ICCV}, 2019.

\bibitem{anwar2020densely}
Saeed Anwar and Nick Barnes.
\newblock Densely residual laplacian super-resolution.
\newblock {\em TPAMI}, 2020.

\bibitem{anwar2019deep}
Saeed Anwar, Salman Khan, and Nick Barnes.
\newblock A deep journey into super-resolution: A survey.
\newblock {\em ACM Computing Surveys}, 2019.

\bibitem{Brooks2019}
Tim Brooks, Ben Mildenhall, Tianfan Xue, Jiawen Chen, Dillon Sharlet, and
  Jonathan~T Barron.
\newblock Unprocessing images for learned raw denoising.
\newblock In {\em CVPR}, 2019.

\bibitem{NLM}
Antoni Buades, Bartomeu Coll, and J-M Morel.
\newblock A non-local algorithm for image denoising.
\newblock In {\em CVPR}, 2005.

\bibitem{MLP}
Harold~C Burger, Christian~J Schuler, and Stefan Harmeling.
\newblock Image denoising: Can plain neural networks compete with {BM3D}?
\newblock In {\em CVPR}, 2012.

\bibitem{chan1998total}
Tony~F Chan and Chiu-Kwong Wong.
\newblock Total variation blind deconvolution.
\newblock {\em TIP}, 1998.

\bibitem{chang2020sadnet}
Meng Chang, Qi Li, Huajun Feng, and Zhihai Xu.
\newblock Spatial-adaptive network for single image denoising.
\newblock In {\em ECCV}, 2020.

\bibitem{charbonnier1994}
Pierre Charbonnier, Laure Blanc-Feraud, Gilles Aubert, and Michel Barlaud.
\newblock Two deterministic half-quadratic regularization algorithms for
  computed imaging.
\newblock In {\em ICIP}, 1994.

\bibitem{Chen2018}
Chen Chen, Qifeng Chen, Jia Xu, and Vladlen Koltun.
\newblock Learning to see in the dark.
\newblock In {\em CVPR}, 2018.

\bibitem{chen2018cascaded}
Yilun Chen, Zhicheng Wang, Yuxiang Peng, Zhiqiang Zhang, Gang Yu, and Jian Sun.
\newblock Cascaded pyramid network for multi-person pose estimation.
\newblock In {\em CVPR}, 2018.

\bibitem{cheng2019spgnet}
Bowen Cheng, Liang-Chieh Chen, Yunchao Wei, Yukun Zhu, Zilong Huang, Jinjun
  Xiong, Thomas~S Huang, Wen-Mei Hwu, and Honghui Shi.
\newblock {SPGNet}: Semantic prediction guidance for scene parsing.
\newblock In {\em ICCV}, 2019.

\bibitem{BM3D}
Kostadin Dabov, Alessandro Foi, Vladimir Katkovnik, and Karen Egiazarian.
\newblock Image denoising by sparse {3-D} transform-domain collaborative
  filtering.
\newblock {\em TIP}, 2007.

\bibitem{dai2019second}
Tao Dai, Jianrui Cai, Yongbing Zhang, Shu-Tao Xia, and Lei Zhang.
\newblock Second-order attention network for single image super-resolution.
\newblock In {\em CVPR}, 2019.

\bibitem{dong2015image}
Chao Dong, Chen~Change Loy, Kaiming He, and Xiaoou Tang.
\newblock Image super-resolution using deep convolutional networks.
\newblock {\em TPAMI}, 2015.

\bibitem{dong2011image}
Weisheng Dong, Lei Zhang, Guangming Shi, and Xiaolin Wu.
\newblock Image deblurring and super-resolution by adaptive sparse domain
  selection and adaptive regularization.
\newblock {\em TIP}, 2011.

\bibitem{farha2019ms}
Yazan~Abu Farha and Jurgen Gall.
\newblock {MS-TCN}: Multi-stage temporal convolutional network for action
  segmentation.
\newblock In {\em CVPR}, 2019.

\bibitem{fu2019dual}
Jun Fu, Jing Liu, Haijie Tian, Yong Li, Yongjun Bao, Zhiwei Fang, and Hanqing
  Lu.
\newblock Dual attention network for scene segmentation.
\newblock In {\em CVPR}, 2019.

\bibitem{fu2017clearing}
Xueyang Fu, Jiabin Huang, Xinghao Ding, Yinghao Liao, and John Paisley.
\newblock Clearing the skies: A deep network architecture for single-image rain
  removal.
\newblock {\em TIP}, 2017.

\bibitem{fu2017removing}
Xueyang Fu, Jiabin Huang, Delu Zeng, Yue Huang, Xinghao Ding, and John Paisley.
\newblock Removing rain from single images via a deep detail network.
\newblock In {\em CVPR}, 2017.

\bibitem{fu2019lightweight}
Xueyang Fu, Borong Liang, Yue Huang, Xinghao Ding, and John Paisley.
\newblock Lightweight pyramid networks for image deraining.
\newblock {\em TNNLS}, 2019.

\bibitem{gao2019dynamic}
Hongyun Gao, Xin Tao, Xiaoyong Shen, and Jiaya Jia.
\newblock Dynamic scene deblurring with parameter selective sharing and nested
  skip connections.
\newblock In {\em CVPR}, 2019.

\bibitem{ghosh2020stacked}
Pallabi Ghosh, Yi Yao, Larry Davis, and Ajay Divakaran.
\newblock Stacked spatio-temporal graph convolutional networks for action
  segmentation.
\newblock In {\em WACV}, 2020.

\bibitem{gong2017motion}
Dong Gong, Jie Yang, Lingqiao Liu, Yanning Zhang, Ian Reid, Chunhua Shen, Anton
  Van Den~Hengel, and Qinfeng Shi.
\newblock From motion blur to motion flow: a deep learning solution for
  removing heterogeneous motion blur.
\newblock In {\em CVPR}, 2017.

\bibitem{CBDNet}
Shi Guo, Zifei Yan, Kai Zhang, Wangmeng Zuo, and Lei Zhang.
\newblock Toward convolutional blind denoising of real photographs.
\newblock In {\em CVPR}, 2019.

\bibitem{he2010single}
Kaiming He, Jian Sun, and Xiaoou Tang.
\newblock Single image haze removal using dark channel prior.
\newblock {\em TPAMI}, 2010.

\bibitem{he2016deep}
Kaiming He, Xiangyu Zhang, Shaoqing Ren, and Jian Sun.
\newblock Deep residual learning for image recognition.
\newblock In {\em CVPR}, 2016.

\bibitem{hu2018gather}
Jie Hu, Li Shen, Samuel Albanie, Gang Sun, and Andrea Vedaldi.
\newblock Gather-excite: Exploiting feature context in convolutional neural
  networks.
\newblock In {\em NeurIPS}, 2018.

\bibitem{hu2019squeeze}
Jie Hu, Li Shen, Samuel Albanie, Gang Sun, and Enhua Wu.
\newblock Squeeze-and-excitation networks.
\newblock {\em IEEE TPAMI}, 2019.

\bibitem{hu2014deblurring}
Zhe Hu, Sunghyun Cho, Jue Wang, and Ming-Hsuan Yang.
\newblock Deblurring low-light images with light streaks.
\newblock In {\em CVPR}, 2014.

\bibitem{huang2017densely}
Gao Huang, Zhuang Liu, Laurens Van Der~Maaten, and Kilian~Q Weinberger.
\newblock Densely connected convolutional networks.
\newblock In {\em CVPR}, 2017.

\bibitem{huang2019ccnet}
Zilong Huang, Xinggang Wang, Lichao Huang, Chang Huang, Yunchao Wei, and Wenyu
  Liu.
\newblock {CCNet}: Criss-cross attention for semantic segmentation.
\newblock In {\em ICCV}, 2019.

\bibitem{hyun2013dynamic}
Tae Hyun~Kim, Byeongjoo Ahn, and Kyoung Mu~Lee.
\newblock Dynamic scene deblurring.
\newblock In {\em ICCV}, 2013.

\bibitem{mspfn2020}
Kui Jiang, Zhongyuan Wang, Peng Yi, Baojin Huang, Yimin Luo, Jiayi Ma, and
  Junjun Jiang.
\newblock Multi-scale progressive fusion network for single image deraining.
\newblock In {\em CVPR}, 2020.

\bibitem{khan2021transformers}
Salman Khan, Muzammal Naseer, Munawar Hayat, Syed~Waqas Zamir, Fahad~Shahbaz
  Khan, and Mubarak Shah.
\newblock Transformers in vision: A survey.
\newblock {\em arXiv:2101.01169}, 2021.

\bibitem{kim2010single}
Kwang~In Kim and Younghee Kwon.
\newblock Single-image super-resolution using sparse regression and natural
  image prior.
\newblock {\em TPAMI}, 2010.

\bibitem{kim2020aindnet}
Yoonsik Kim, Jae~Woong Soh, Gu~Yong Park, and Nam~Ik Cho.
\newblock Transfer learning from synthetic to real-noise denoising with
  adaptive instance normalization.
\newblock In {\em CVPR}, 2020.

\bibitem{kingma2014adam}
Diederik~P Kingma and Jimmy Ba.
\newblock Adam: A method for stochastic optimization.
\newblock {\em arXiv:1412.6980}, 2014.

\bibitem{deblurgan}
Orest Kupyn, Volodymyr Budzan, Mykola Mykhailych, Dmytro Mishkin, and
  Ji{\v{r}}{\'\i} Matas.
\newblock {DeblurGAN}: Blind motion deblurring using conditional adversarial
  networks.
\newblock In {\em CVPR}, 2018.

\bibitem{deblurganv2}
Orest Kupyn, Tetiana Martyniuk, Junru Wu, and Zhangyang Wang.
\newblock {DeblurGAN-v2}: Deblurring (orders-of-magnitude) faster and better.
\newblock In {\em ICCV}, 2019.

\bibitem{SRResNet}
Christian Ledig, Lucas Theis, Ferenc Husz{\'a}r, Jose Caballero, Andrew
  Cunningham, Alejandro Acosta, Andrew Aitken, Alykhan Tejani, Johannes Totz,
  Zehan Wang, et~al.
\newblock Photo-realistic single image super-resolution using a generative
  adversarial network.
\newblock In {\em CVPR}, 2017.

\bibitem{li2020ms}
Shi-Jie Li, Yazan AbuFarha, Yun Liu, Ming-Ming Cheng, and Juergen Gall.
\newblock {MS-TCN++}: Multi-stage temporal convolutional network for action
  segmentation.
\newblock {\em TPAMI}, 2020.

\bibitem{li2019rethinking}
Wenbo Li, Zhicheng Wang, Binyi Yin, Qixiang Peng, Yuming Du, Tianzi Xiao, Gang
  Yu, Hongtao Lu, Yichen Wei, and Jian Sun.
\newblock Rethinking on multi-stage networks for human pose estimation.
\newblock {\em arXiv:1901.00148}, 2019.

\bibitem{li2018recurrent}
Xia Li, Jianlong Wu, Zhouchen Lin, Hong Liu, and Hongbin Zha.
\newblock Recurrent squeeze-and-excitation context aggregation net for single
  image deraining.
\newblock In {\em ECCV}, 2018.

\bibitem{li2016rain}
Yu Li, Robby~T Tan, Xiaojie Guo, Jiangbo Lu, and Michael~S Brown.
\newblock Rain streak removal using layer priors.
\newblock In {\em CVPR}, 2016.

\bibitem{liu2015gap}
Wei Liu, Andrew Rabinovich, and Alexander~C Berg.
\newblock {ParseNet}: Looking wider to see better.
\newblock {\em arXiv:1506.04579}, 2015.

\bibitem{loshchilov2016sgdr}
Ilya Loshchilov and Frank Hutter.
\newblock {SGDR}: Stochastic gradient descent with warm restarts.
\newblock In {\em ICLR}, 2017.

\bibitem{luo2015removing}
Yu Luo, Yong Xu, and Hui Ji.
\newblock Removing rain from a single image via discriminative sparse coding.
\newblock In {\em ICCV}, 2015.

\bibitem{mairal2007sparse}
Julien Mairal, Michael Elad, and Guillermo Sapiro.
\newblock Sparse representation for color image restoration.
\newblock {\em TIP}, 2007.

\bibitem{gopro2017}
Seungjun Nah, Tae Hyun~Kim, and Kyoung Mu~Lee.
\newblock Deep multi-scale convolutional neural network for dynamic scene
  deblurring.
\newblock In {\em CVPR}, 2017.

\bibitem{newell2016stacked}
Alejandro Newell, Kaiyu Yang, and Jia Deng.
\newblock Stacked hourglass networks for human pose estimation.
\newblock In {\em ECCV}, 2016.

\bibitem{Odena2016}
Augustus Odena, Vincent Dumoulin, and Chris Olah.
\newblock Deconvolution and checkerboard artifacts.
\newblock {\em Distill}, 2016.

\bibitem{pan2016blind}
Jinshan Pan, Deqing Sun, Hanspeter Pfister, and Ming-Hsuan Yang.
\newblock Blind image deblurring using dark channel prior.
\newblock In {\em CVPR}, 2016.

\bibitem{pan2020exploiting}
Xingang Pan, Xiaohang Zhan, Bo Dai, Dahua Lin, Chen~Change Loy, and Ping Luo.
\newblock Exploiting deep generative prior for versatile image restoration and
  manipulation.
\newblock In {\em ECCV}, 2020.

\bibitem{mtrnn2020}
Dongwon Park, Dong~Un Kang, Jisoo Kim, and Se~Young Chun.
\newblock Multi-temporal recurrent neural networks for progressive non-uniform
  single image deblurring with incremental temporal training.
\newblock In {\em ECCV}, 2020.

\bibitem{perona1990scale}
Pietro Perona and Jitendra Malik.
\newblock Scale-space and edge detection using anisotropic diffusion.
\newblock {\em TPAMI}, 1990.

\bibitem{dnd}
Tobias Plotz and Stefan Roth.
\newblock Benchmarking denoising algorithms with real photographs.
\newblock In {\em CVPR}, 2017.

\bibitem{purohit2020region}
Kuldeep Purohit and AN Rajagopalan.
\newblock Region-adaptive dense network for efficient motion deblurring.
\newblock In {\em AAAI}, 2020.

\bibitem{qian2018attentive}
Rui Qian, Robby~T Tan, Wenhan Yang, Jiajun Su, and Jiaying Liu.
\newblock Attentive generative adversarial network for raindrop removal from a
  single image.
\newblock In {\em CVPR}, 2018.

\bibitem{ren2019progressive}
Dongwei Ren, Wangmeng Zuo, Qinghua Hu, Pengfei Zhu, and Deyu Meng.
\newblock Progressive image deraining networks: A better and simpler baseline.
\newblock In {\em CVPR}, 2019.

\bibitem{rim_2020_realblur}
Jaesung Rim, Haeyun Lee, Jucheol Won, and Sunghyun Cho.
\newblock Real-world blur dataset for learning and benchmarking deblurring
  algorithms.
\newblock In {\em ECCV}, 2020.

\bibitem{ronneberger2015unet}
Olaf Ronneberger, Philipp Fischer, and Thomas Brox.
\newblock {U-Net:} convolutional networks for biomedical image segmentation.
\newblock In {\em MICCAI}, 2015.

\bibitem{roth2005fields}
Stefan Roth and Michael~J Black.
\newblock Fields of experts: A framework for learning image priors.
\newblock In {\em CVPR}, 2005.

\bibitem{rudin1992nonlinear}
Leonid~I Rudin, Stanley Osher, and Emad Fatemi.
\newblock Nonlinear total variation based noise removal algorithms.
\newblock {\em Physica D: nonlinear phenomena}, 1992.

\bibitem{shan2008high}
Qi Shan, Jiaya Jia, and Aseem Agarwala.
\newblock High-quality motion deblurring from a single image.
\newblock {\em ToG}, 2008.

\bibitem{shen2019human}
Ziyi Shen, Wenguan Wang, Xiankai Lu, Jianbing Shen, Haibin Ling, Tingfa Xu, and
  Ling Shao.
\newblock Human-aware motion deblurring.
\newblock In {\em ICCV}, 2019.

\bibitem{Maitreya2020}
Maitreya Suin, Kuldeep Purohit, and A.~N. Rajagopalan.
\newblock Spatially-attentive patch-hierarchical network for adaptive motion
  deblurring.
\newblock In {\em CVPR}, 2020.

\bibitem{tao2018scale}
Xin Tao, Hongyun Gao, Xiaoyong Shen, Jue Wang, and Jiaya Jia.
\newblock Scale-recurrent network for deep image deblurring.
\newblock In {\em CVPR}, 2018.

\bibitem{tian2020deep}
Chunwei Tian, Lunke Fei, Wenxian Zheng, Yong Xu, Wangmeng Zuo, and Chia-Wen
  Lin.
\newblock Deep learning on image denoising: An overview.
\newblock {\em Neural Networks}, 2020.

\bibitem{tong2017image}
Tong Tong, Gen Li, Xiejie Liu, and Qinquan Gao.
\newblock Image super-resolution using dense skip connections.
\newblock In {\em ICCV}, 2017.

\bibitem{wang2018non}
Xiaolong Wang, Ross Girshick, Abhinav Gupta, and Kaiming He.
\newblock Non-local neural networks.
\newblock In {\em CVPR}, 2018.

\bibitem{wang2018esrgan}
Xintao Wang, Ke Yu, Shixiang Wu, Jinjin Gu, Yihao Liu, Chao Dong, Yu Qiao, and
  Chen Change~Loy.
\newblock {ESRGAN:} enhanced super-resolution generative adversarial networks.
\newblock In {\em ECCVW}, 2018.

\bibitem{Wang2004ssim}
Zhou Wang, A.~C. Bovik, H.~R. Sheikh, and E.~P. Simoncelli.
\newblock Image quality assessment: from error visibility to structural
  similarity.
\newblock {\em TIP}, 2004.

\bibitem{wei2019semi}
Wei Wei, Deyu Meng, Qian Zhao, Zongben Xu, and Ying Wu.
\newblock Semi-supervised transfer learning for image rain removal.
\newblock In {\em CVPR}, 2019.

\bibitem{whyte2012non}
Oliver Whyte, Josef Sivic, Andrew Zisserman, and Jean Ponce.
\newblock Non-uniform deblurring for shaken images.
\newblock {\em IJCV}, 2012.

\bibitem{woo2018cbam}
Sanghyun Woo, Jongchan Park, Joon-Young Lee, and In So~Kweon.
\newblock Cbam: Convolutional block attention module.
\newblock In {\em ECCV}, 2018.

\bibitem{xu2013unnatural}
Li Xu, Shicheng Zheng, and Jiaya Jia.
\newblock Unnatural l0 sparse representation for natural image deblurring.
\newblock In {\em CVPR}, 2013.

\bibitem{yang2017deep}
Wenhan Yang, Robby~T Tan, Jiashi Feng, Jiaying Liu, Zongming Guo, and Shuicheng
  Yan.
\newblock Deep joint rain detection and removal from a single image.
\newblock In {\em CVPR}, 2017.

\bibitem{yasarla2019uncertainty}
Rajeev Yasarla and Vishal~M Patel.
\newblock Uncertainty guided multi-scale residual learning-using a cycle
  spinning cnn for single image de-raining.
\newblock In {\em CVPR}, 2019.

\bibitem{yu2015multi}
Fisher Yu and Vladlen Koltun.
\newblock Multi-scale context aggregation by dilated convolutions.
\newblock In {\em ICLR}, 2016.

\bibitem{VDN}
Zongsheng Yue, Hongwei Yong, Qian Zhao, Deyu Meng, and Lei Zhang.
\newblock Variational denoising network: Toward blind noise modeling and
  removal.
\newblock In {\em NeurIPS}, 2019.

\bibitem{yue2020danet}
Zongsheng Yue, Qian Zhao, Lei Zhang, and Deyu Meng.
\newblock Dual adversarial network: Toward real-world noise removal and noise
  generation.
\newblock In {\em ECCV}, 2020.

\bibitem{zamir2020cycleisp}
Syed~Waqas Zamir, Aditya Arora, Salman Khan, Munawar Hayat, Fahad~Shahbaz Khan,
  Ming-Hsuan Yang, and Ling Shao.
\newblock {CycleISP}: Real image restoration via improved data synthesis.
\newblock In {\em CVPR}, 2020.

\bibitem{zamir2020mirnet}
Syed~Waqas Zamir, Aditya Arora, Salman Khan, Munawar Hayat, Fahad~Shahbaz Khan,
  Ming-Hsuan Yang, and Ling Shao.
\newblock Learning enriched features for real image restoration and
  enhancement.
\newblock In {\em ECCV}, 2020.

\bibitem{dmphn2019}
Hongguang Zhang, Yuchao Dai, Hongdong Li, and Piotr Koniusz.
\newblock Deep stacked hierarchical multi-patch network for image deblurring.
\newblock In {\em CVPR}, 2019.

\bibitem{zhang2018density}
He Zhang and Vishal~M Patel.
\newblock Density-aware single image de-raining using a multi-stream dense
  network.
\newblock In {\em CVPR}, 2018.

\bibitem{zhang2019image}
He Zhang, Vishwanath Sindagi, and Vishal~M Patel.
\newblock Image de-raining using a conditional generative adversarial network.
\newblock {\em TCSVT}, 2019.

\bibitem{zhang2018dynamic}
Jiawei Zhang, Jinshan Pan, Jimmy Ren, Yibing Song, Linchao Bao, Rynson~WH Lau,
  and Ming-Hsuan Yang.
\newblock Dynamic scene deblurring using spatially variant recurrent neural
  networks.
\newblock In {\em CVPR}, 2018.

\bibitem{zhang2020dbgan}
Kaihao Zhang, Wenhan Luo, Yiran Zhong, Lin Ma, Bjorn Stenger, Wei Liu, and
  Hongdong Li.
\newblock Deblurring by realistic blurring.
\newblock In {\em CVPR}, 2020.

\bibitem{DnCNN}
Kai Zhang, Wangmeng Zuo, Yunjin Chen, Deyu Meng, and Lei Zhang.
\newblock Beyond a gaussian denoiser: Residual learning of deep cnn for image
  denoising.
\newblock {\em TIP}, 2017.

\bibitem{zhang2017learning}
Kai Zhang, Wangmeng Zuo, Shuhang Gu, and Lei Zhang.
\newblock Learning deep cnn denoiser prior for image restoration.
\newblock In {\em CVPR}, 2017.

\bibitem{RCAN}
Yulun Zhang, Kunpeng Li, Kai Li, Lichen Wang, Bineng Zhong, and Yun Fu.
\newblock Image super-resolution using very deep residual channel attention
  networks.
\newblock In {\em ECCV}, 2018.

\bibitem{zhang2019residual}
Yulun Zhang, Kunpeng Li, Kai Li, Bineng Zhong, and Yun Fu.
\newblock Residual non-local attention networks for image restoration.
\newblock In {\em ICLR}, 2019.

\bibitem{zhang2020rdn}
Yulun Zhang, Yapeng Tian, Yu Kong, Bineng Zhong, and Yun Fu.
\newblock Residual dense network for image restoration.
\newblock {\em TPAMI}, 2020.

\bibitem{zhao2018psanet}
Hengshuang Zhao, Yi Zhang, Shu Liu, Jianping Shi, Chen Change~Loy, Dahua Lin,
  and Jiaya Jia.
\newblock Psanet: Point-wise spatial attention network for scene parsing.
\newblock In {\em ECCV}, 2018.

\bibitem{zheng2019residual}
Yupei Zheng, Xin Yu, Miaomiao Liu, and Shunli Zhang.
\newblock Residual multiscale based single image deraining.
\newblock In {\em BMVC}, 2019.

\bibitem{zhu1997prior}
Song~Chun Zhu and David Mumford.
\newblock Prior learning and gibbs reaction-diffusion.
\newblock {\em TPAMI}, 1997.

\end{thebibliography}
}

\end{document}